\documentclass[10pt,journal,compsoc]{IEEEtran}

\ifCLASSINFOpdf
\else
\fi

\usepackage{cite}
\usepackage{amsmath,amssymb,amsfonts}
\usepackage{algorithmic}
\usepackage[linesnumbered,algoruled,boxed,lined]{algorithm2e}
\usepackage{graphicx}
\usepackage{textcomp}
\usepackage{xcolor}
\usepackage{multirow}
\usepackage{bm}
\usepackage{booktabs}
\usepackage{ntheorem}
\usepackage{tabularx}
\usepackage[caption=false, font=footnotesize, labelfont=sf, textfont=sf]{subfig}

\theoremseparator{.}
\newskip\theorempreskipamount
\newskip\theorempostskipamount

\newtheorem{definition}{Definition}
\newtheorem{lemma}{Lemma}
\newtheorem{proposition}{Proposition}

\theorembodyfont{}

\newenvironment{proof}{{\noindent\it Proof. }}{\hfill $\blacksquare$\par}

\newcommand\undermat[2]{%
  \makebox[-3pt][l]{$\smash{\underbrace{\phantom{%
    \begin{matrix}#2\end{matrix}}}_{\text{$#1$}}}$}#2}

\allowdisplaybreaks[4]

\begin{document}

\title{Multi-Objective Optimization for UAV Swarm-Assisted IoT with Virtual Antenna Arrays\\
}

\author{Jiahui Li,~\IEEEmembership{Student Member,~IEEE,}
	Geng Sun*,~\IEEEmembership{Member,~IEEE,}
	Lingjie Duan,~\IEEEmembership{Senior Member,~IEEE,}
	and Qingqing Wu,~\IEEEmembership{Senior Member,~IEEE}
	
	\thanks{
 
	\par Jiahui Li is with the College of Computer Science and Technology, Jilin University, Changchun 130012, China, and also with Pillar of Engineering Systems and Design, Singapore University of Technology and Design, Singapore 487372 (e-mail: lijiahui0803@foxmail.com).

	\par Geng Sun is with College of Computer Science and Technology, Jilin University, Changchun 130012, China, and also with Key Laboratory of Symbolic Computation and Knowledge Engineering of Ministry of Education, Jilin University, Changchun 130012, China (e-mail: sungeng@jlu.edu.cn).
	
	\par Lingjie Duan is with Pillar of Engineering Systems and Design, Singapore University of Technology and Design, Singapore 487372 (e-mail: lingjie\_duan@sutd.edu.sg).
	
	\par Qingqing Wu is with the Department of Electronic Engineering, Shanghai Jiao Tong University, Shanghai 200240, China (e-mail: qingqingwu@sjtu.edu.cn).

	}
	}

	


\IEEEtitleabstractindextext{%
\begin{abstract}
Unmanned aerial vehicle (UAV) network is a promising technology for assisting Internet-of-Things (IoT), where a UAV can use its limited service coverage to harvest and disseminate data from IoT devices with low transmission abilities. The existing UAV-assisted data harvesting and dissemination schemes largely require UAVs to frequently fly between the IoTs and access points, resulting in extra energy and time costs. To reduce both energy and time costs, a key way is to enhance the transmission performance of IoT and UAVs. In this work, we introduce collaborative beamforming into IoTs and UAVs simultaneously to achieve energy and time-efficient data harvesting and dissemination from multiple IoT clusters to remote base stations (BSs). Except for reducing these costs, another non-ignorable threat lies in the existence of the potential eavesdroppers, whereas the handling of eavesdroppers often increases the energy and time costs, resulting in a conflict with the minimization of the costs. Moreover, the importance of these goals may vary relatively in different applications. Thus, we formulate a multi-objective optimization problem (MOP) to simultaneously minimize the mission completion time, signal strength towards the eavesdropper, and total energy cost of the UAVs. We prove that the formulated MOP is an NP-hard, mixed-variable optimization, and large-scale optimization problem. Thus, we propose a swarm intelligence-based algorithm to find a set of candidate solutions with different trade-offs which can meet various requirements in a low computational complexity. We also show that swarm intelligence methods need to enhance solution initialization, solution update, and algorithm parameter update phases when dealing with mixed-variable optimization and large-scale problems. Simulation results demonstrate the proposed algorithm outperforms state-of-the-art swarm intelligence algorithms and also show that the proposed method can reduce time and energy costs significantly compared with the benchmark strategies based on multi-hop and long-range flight.
\end{abstract}

}

\maketitle
%

\IEEEdisplaynontitleabstractindextext
\IEEEpeerreviewmaketitle

\IEEEraisesectionheading{
%
%
\section{Introduction} 
\label{sec:introduction}
}

\IEEEPARstart{W}{ith} the manufacturing improvements and cost reduction of the Internet-of-Things (IoT), massive devices are connected to the networks and various applications provide critical support to people's daily life. IoT devices, e.g., wireless sensor~\cite{Luong2016}, are often installed in hard-to-reach locations, and thus it is difficult to upload the sensed data to the remote access points. Recently, unmanned aerial vehicle (UAV)-enabled data harvesting and dissemination in IoT systems is a promising technology. Specifically, the UAVs can fly to the unreachable area for data harvesting and then disseminate the collected data to the remote access points via their high flexibility and maneuverability. Moreover, due to their high flight altitude, the UAVs can obtain a higher line-of-sight (LoS) probability, thereby enhancing the channel conditions for data transmissions~\cite{Wu2021,Jiang2023}. In addition, UAVs can be deployed rapidly~\cite{Wang2021,Wang2023}, such that improving the coverage and quality of UAV-enabled data harvesting and dissemination networks~\cite{Zhan2018}.

\par
\begin{figure}
    \centering
	  \subfloat[]{
       \includegraphics[width=0.95\linewidth]{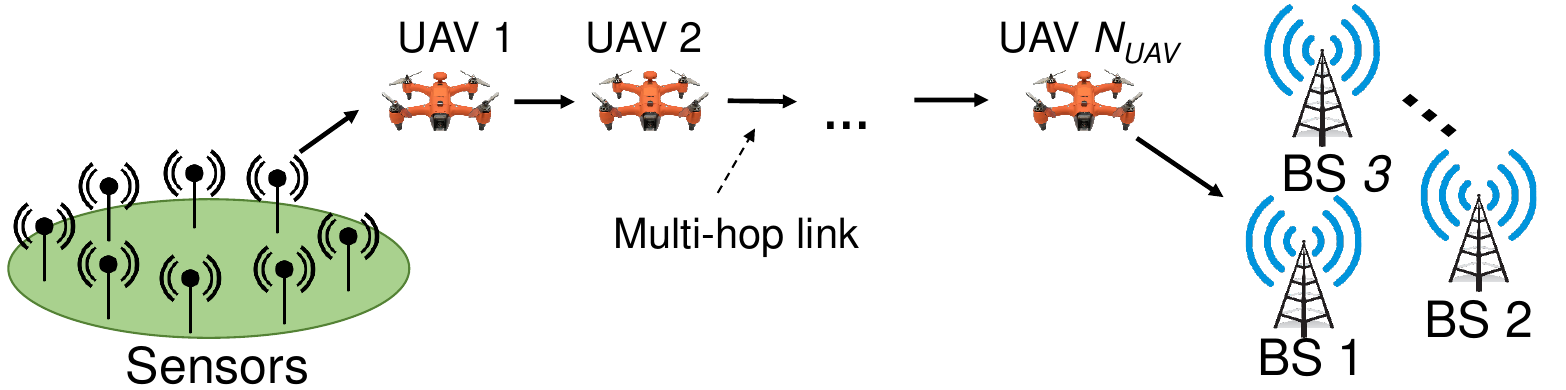}\label{fig:benchmark1}}
    \\
    \centering
	  \subfloat[]{
        \includegraphics[width=0.95\linewidth]{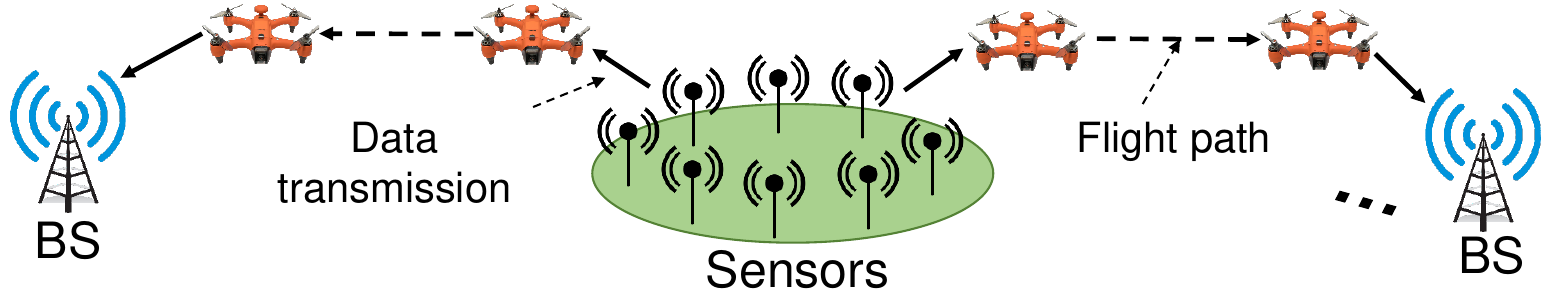}\label{fig:benchmark2}}
    \\
	\caption{UAV-assistant data harvesting and dissemination strategies. (a) Constructing a multi-hop flying ad-hoc network. (b) Flying between the sensors and BSs.}
	\label{fig:benchmark}
\end{figure}

 However, the UAVs only have limited service coverage due to the constrained energy and antenna with insufficient transmit power and directivity~\cite{Zeng2019a,Gong_2023}, likewise, the sensors also have a low transmission ability, which causes high energy and time costs for data harvesting and dissemination. Concretely, for accomplishing the data harvesting and dissemination tasks, the existing works largely require the UAVs to frequently fly between the sensors and access points. For example, as shown in Fig.~\ref{fig:benchmark}, some methods require the UAVs fly between the sensors and access points to construct a UAV-enabled multi-hop link for transmitting data (e.g.,~\cite{Kim2020, Ranjha2020, Banerjee2022,Hong2019}), moreover, some methods adopt a UAV to collect data from the sensors and then fly close to access points for data dissemination (e.g.,~\cite{Zhan2018, Hu2021, You2019, Xu2021}). However, long-range or frequent flights will undoubtedly increase the time and energy costs of the UAVs. Hence, it is necessary to enhance the transmission abilities of both UAVs and sensors for reducing these frequent flights, thereby increasing the service quality and coverage of UAVs.

%

\par In this work, we aim to introduce collaborative beamforming (CB) technology to enhance the communication performance of UAVs and sensors simultaneously. Specifically, multiple array elements at UAVs or sensors can form a virtual antenna array. Then, the virtual antenna array synchronizes among array elements to transmit, thereby getting constructive interference at the location of the receiver~\cite{Sun2022}. Benefiting from the virtual antenna gain, the virtual antenna array can achieve $N^2$ fold gain in the received power at the destination via $N$ array elements~\cite{Jayaprakasam2017}. Thus, CB can make up the insufficient transmission gains of UAVs and sensors without changing the equipped devices, thereby avoiding the over-frequent trajectory changes of UAVs. Simultaneously, since the transmission ranges and gains are enhanced, the probability of being eavesdropped on may also increase. In this case, we can carefully design the virtual antenna array to get destructive interference at the location of the eavesdropper for achieving physical layer security.

\par Such CB-based systems are applicable in many real cases. For instance, in some agricultural scenarios, sensors are distributed in distant farmlands to detect soil moisture, humidity, temperature, and sap flow~\cite{Pei2022IoT}~\cite{Zhou2022sleeping}. When the data needs to be uploaded to a remote data center for data fusion, the data can be relayed by the UAVs which are dispatched for spreading pesticides. In this case, the CB-based method allows UAVs to remain in their original regions for data forwarding tasks, which facilitates UAVs to continue their original tasks after the relay task is completed. Since UAVs do not need to fly close to BSs or bypass eavesdropper, the corresponding flight time and energy costs of UAVs can be significantly saved. 


\par However, the performance of the considered system may be degraded by some handicaps, e.g., the randomly distributed sensors and UAVs may damage the beam patterns of the sensor-enabled ground virtual antenna arrays (GVAAs) and UAV-enabled aerial virtual antenna arrays (AVAAs). Accordingly, we need prudently optimize the selection of the sensors and positions of the UAVs for CB. Moreover, in CB, the excitation current weights (also known as beamforming coefficients) of array elements in both the GVAAs and AVAAs, are another key factor that affects the performance of CB. On the other hand, such a complex system also requires appropriate operational strategies. Specifically, we need to select the proper aerial UAVs of the swarm for receiving the harvested data from the ground sensors since these UAVs are both receivers and array elements. Moreover, when the data need to be disseminated to multiple access points, e.g., base stations (BSs), the order to access these BSs needs to be carefully designed since it affects the energy cost. Thus, these issues above need to be carefully designed and optimized for achieving the considered goals.

\par Nevertheless, finding a reasonable trade-off between time cost, energy cost, and secure performance of the considered system is a challenging task. On the one hand, the time cost for completing the data harvesting and dissemination mission in our system is related to the speed control of UAVs which also affects the energy cost, which means that the time and energy costs are difficult to be reduced simultaneously. On the other hand, the improvement of secure performance may degrade the directivity of the virtual antenna array, resulting in extra time costs for transmission~\cite{Li2021}. Thus, these three goals are in conflict with each other and are hard to be balanced. Additionally, under different scenarios, their importance may vary relatively, which means some existing optimization methods in literature (e.g.,~\cite{Hua2019,Cai2020}) that only consider a single optimization goal and add others as constraints are inapplicable. Thus, it requires a new multi-objective optimization analysis which is different from the literature. We summarize the key novelty and main contributions of this paper as follows.

\begin{itemize}

  \item \textit{CB-based UAV-assisted Data Harvesting and Dissemination System for IoT}: We introduce CB into the UAV-assisted data harvesting and dissemination system for IoT, where the sensors and UAVs are formed as virtual antenna arrays to enhance the transmission performance simultaneously. This system enhances the insufficient transmission abilities of UAVs and sensors without changing the equipped devices, thereby reducing the frequent and over-range flights of UAVs for time and energy efficiencies. Note that to the best of our knowledge, such a joint optimization of CB in both ground and air communications has not been investigated yet in the literature.

  \item \textit{Multi-objective Optimization Formulation}: We find that the time cost, energy cost, and secure performance of the considered system are in conflict with each other. Accordingly, we adopt a multi-objective optimization method to balance these goals, in which a multi-objective optimization problem (MOP) is formulated to simultaneously reduce the data transmission time, suppress the signal intensity of the eavesdropper, and save energy cost of the UAV swarm. This problem is difficult and we prove it to be an NP-hard, mixed-variable, and large-scale problem.

  \item \textit{New Swarm Intelligence Method}: We propose a swarm intelligence-based algorithm, namely, enhanced multi-objective salp swarm algorithm (EMSSA), to solve the formulated MOP. This algorithm is able to find a set of candidate solutions with different trade-offs which can meet various requirements in a low computational complexity. Moreover, the algorithm is able to handle four different types of decision variables via the swarm intelligent manner and provides effective solutions to some physical constraints.

  \item \textit{Simulation and Performance Evaluation}: Extensive simulation results demonstrate the proposed EMSSA outmatches various multi-objective swarm intelligence algorithms. Moreover, we find that the proposed method can reduce time and energy costs significantly compared with some benchmark strategies which require the UAVs to frequently fly (such as the UAVs constructing a multi-hop link or directly flying between sensors and access points).

\end{itemize}

\par The rest of this paper is arranged as follows. Section \ref{sec:related_works} reviews the related research activities. Section \ref{sec:models_and_preliminaries} presents the models. Section \ref{sec:problem_formulation_and_analysis} formulates the MOP. Multi-objective swarm intelligence and the proposed EMSSA are introduced in Sections \ref{sec:multi_objective_optimization_and_swarm_intelligence} and \ref{sec:EMSSA}, respectively. Simulation results are presented in Section \ref{sec:simulation_results_and_analysis}. Finally, the paper is concluded in Section \ref{sec:conclusion}.

\section{Related Work} 
\label{sec:related_works}

\par In the existing works of wireless communications and networks, there are some studies of UAV-assisted data harvesting and dissemination in IoT, collaborative beamforming and multi-objective optimization, and we briefly introduce them in this section.

%
%
\subsection{UAV-assisted Data Harvesting and Dissemination in IoT}

\par Most literature about the UAV-assisted data harvesting and dissemination methods in IoT focused on designing the energy-efficient trajectory or deployment schemes of UAVs to improve efficiency.

\par Some existing works constructed a multi-hop link by using UAVs between the sensors in IoT and access points for data transmissions. In reference~\cite{Kim2020}, the authors studied an energy-efficient and cooperative multi-hop relay scheme for UAVs to harvest data from IoT and formulated an optimal multi-hop transmission scheduling problem to minimize the power consumption of UAVs. Ranjha et al.~\cite{Ranjha2020} used multi-hop UAV relay links to deliver short ultra-reliable and low-latency (URLLC) instruction packets between ground IoT devices. The authors in reference~\cite{Choudhury2021} studied a UAV-enabled multi-hop network to collect information from IoT devices. Then, the collected information was delivered to the nearby BS, in which a scheduling policy for the age of information (AoI) minimization was proposed.

\par Several existing works that used UAVs flying between IoTs and access points for data harvesting and dissemination. For example, Zhan et al. \cite{Zhan2018} investigated a UAV-enabled IoT communication system in which a UAV is dispatched to harvest the sensed data from the distributed sensors, and they formulated a complex optimization problem to design the UAV's flight path subject to the motion constraints and then proposed an efficient suboptimal method. In \cite{Hu2021}, the UAV-assisted wireless powered IoT system was investigated, in which a UAV takes off from a data center, flies to each of the ground sensors to transfer energy and collect data, and then returns to the data center, where the AoI of the data acquired from all ground sensors is minimized. You et al. \cite{You2019} considered a UAV-enabled IoT system where a UAV is dispatched to harvest data from the sensors, in which the minimum average data collection rate from sensors is maximized by jointly optimizing the UAV operation and three-dimensional (3D) trajectory under the Rician fading channel. However, these methods lack efficient approaches to handle the potential eavesdroppers. Wang et al. \cite{Wang2020} proposed a UAV-assisted IoT network, in which they minimized the maximum energy consumption of all IoT devices by designing the UAV trajectory and schedule while ensuring the reliability of data collection and required 3D positioning performance. Wang et al. \cite{Wang2019} considered a novel power allocation method for a UAV swarm-enabled network to achieve physical layer security. Dang-Ngoc et al. \cite{DangNgoc2022} considered a UAV swarm relaying information from a terrestrial base station to a distant mobile user and simultaneously generating friendly jamming signals to an eavesdropper.

\par We sum up our differences from these works as follows. First, most works have not considered reconfiguring the IoT network (e.g., performing CB) to facilitate its cooperation with the UAV swarm. Second, these works did not consider using CB technology which may obtain substantial energy and time reduction of UAVs to enhance the transmission ability of UAV-assisted data harvesting and dissemination system. Third, the energy and time costs, and secure performance are in conflict with each other and hard to be balanced, none of the existing works considered a multi-objective optimization approach to optimize them simultaneously and find various optimal trade-offs for meeting different requirements.


%
%
\subsection{Collaborative Beamforming}

\par Many works used CB to improve the transmission abilities of various communication systems. In \cite{Jayaprakasam2017}, the authors detailed the benefits and promising applications of CB in wireless sensor networks. Sun et al.~\cite{Sun2021a} utilized the randomly distributed sensors to form a virtual antenna array by using CB in wireless sensor networks, in which the authors formulated a hybrid optimization problem to reduce the maximum sidelobe level (SLL) and used the centralized and consensus-based distributed CB strategies to solve the problem. However, this method only considered the optimization of SLL while ignoring the mainlobe performance, which may result in a lower transmission rate. Moreover, Mozaffari et al.~\cite{Mozaffari2019} constructed a UAV-enabled linear antenna array and aimed to minimize the wireless transmission and operation times of the UAVs. Li et al. \cite{Li2021} studied a physical layer security communication in UAV networks by using CB, in which a multi-objective optimization approach is proposed to improve the secrecy rate, maximum SLL, and energy consumption of the UAVs simultaneously. However, the above two methods only considered the simplified and ideal LoS channel model, which may not be suitable for practical systems with fading. In addition, the authors in \cite{Mohanti2019} proposed a complete algorithmic framework and system implementation of CB on a set of UAVs, in which some experimental results are reported to demonstrate the practice feasibility of the CB method in UAV networks and beyond. Jung et al. \cite{Jung2022} examined the security energy efficiency of the CB-based physical layer security technique, where a swarm of UAVs randomly changes their locations in a distributed manner. Despite the effectiveness and potential benefits of CB, to the best of our knowledge, no work has been carried out to explore a joint CB optimization in both ground and air networks. This optimization is a challenging task since there are a huge number of mixed decision variables that need to be determined.

%
%
\subsection{Multi-objective Optimization}

\par There are mainly two types of methods to handle MOPs. The first one is to transfer the multiple objectives to a single objective by using the weighting method~\cite{Yu2021weight1, Khan2021weight2,Khan2022weight3}. However, the weights among different objectives are difficult to be determined. Moreover, this method may reduce the solution space, e.g., the linear weighting method can be equivalent to multi-objective optimization algorithms only when the Pareto front of the problem is convex. Another feasible method is to use multi-objective optimization algorithms to optimize all objectives directly, which can obtain a set of Pareto solutions for being chosen by the decision-maker. For instance, aiming to determine the optimal UAV task assignment, Song et al. \cite{Song2022constraint} derived exact Pareto optimal solutions using the $\epsilon$-constraint method and found approximate Pareto solutions via an approximate two-phase approach. However, these approximate method~\cite{Hashir2021nonconvex} cannot deal with some complex decision variables, e.g., constraint programming problems.

\par Multi-objective swarm intelligence algorithms introduce Pareto dominant and congestion management to find a set of candidate solutions with different trade-offs of the MOPs. For instance, Qiu et al.~\cite{Qiu2020modified} introduced and modified multi-objective pigeon-inspired optimization to coordinate UAVs to fly in a stable formation under complex environments. Some state-of-the-art algorithms have been enhanced and adopted in network optimization, e.g.,~\cite{Hashim2019LTE},~\cite{Chourasia2021PSO} and~\cite{Chen2019CRNs}. However, classic swarm intelligence is proposed for dealing with problems with only one type of decision variable, e.g., problems with purely continuous~\cite{Mirjalili2017a}, binary~\cite{Faris2018}, or order~\cite{Mahi2015} decision variables. As such, classic swarm intelligence cannot directly control our system’s key parameters in the forms of binary, continuous, integer, and order types simultaneously. Thus, we aim to propose a novel multi-objective swarm intelligence algorithm that can handle multiple types of decision variables simultaneously in this work.


%
%
\section{Models and Preliminaries} 
\label{sec:models_and_preliminaries}

\par In this section, we present the models and preliminaries used in this work. Specifically, we first present the network models and then present the energy cost model of UAVs.

%
%
\subsection{Network Model}
\label{ssec:data_harvesting_and_dissemination_model}

%
%
\begin{figure}
  \centering
  \includegraphics[width=3.5in]{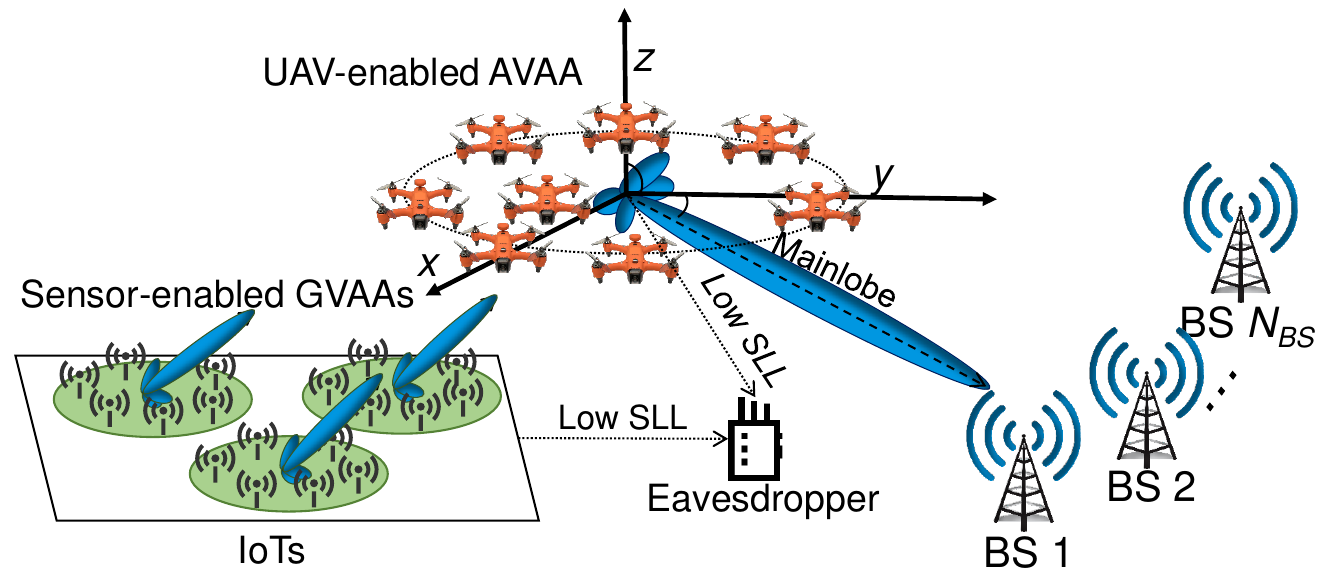}
  \caption{Sketch map of a UAV-assisted data harvesting and dissemination system in IoT based on CB.}
  \label{fig:scenario-model}
\end{figure}

\par As shown in Fig. \ref{fig:scenario-model}, a UAV-assisted data harvesting and dissemination system for serving multiple IoT clusters is considered, in which a monitor area $A_m$ contains massive sensors. Due to the geographic distributions, long communication distance or obstruction can spontaneously divide the sensors into $N_{IoT}$ IoT clusters denoted as $\mathcal{V} \triangleq \{h | W_1, W_2,..., W_{N_{IoT}}\}$. In each IoT cluster, there are a set of sensors for sensing and monitoring, where the sensors of the IoT cluster $h$ can be denoted as $\mathcal{W}_h \triangleq \{ i | 1, 2, \dots, N_{SN}\}$. Due to the insufficient transmit power of sensors and complex terrestrial network environments, it is difficult for the sensors to transmit data to remote BSs directly. In this case, a set of UAVs denoted as $\mathcal{U} \triangleq \{j | 1, 2, \dots, N_{UAV}\}$ are dispatched as a swarm to harvest and disseminate the data from the IoT clusters to a set of remote BSs which is denoted as $\mathcal{B} \triangleq \{k| 1, 2, \dots, N_{BS}\}$, while an eavesdropper denoted as $E$ aims to intercept both the ground-to-air (G2A) and air-to-ground (A2G) communications. In this work, each sensor or UAV is assumed to be equipped with a single omnidirectional antenna, and the position of eavesdropper $E$ can be detected by an optical camera or a synthetic aperture radar equipped on the UAVs.

\par In a certain mission cycle, the data sensed by the sensors in each IoT cluster is aggregated and needs to be transmitted to multiple BSs for the purposes of data backup and data security. Due to the long-range transmission distance between these IoT clusters and BSs, the direct transmissions between them are infeasible, and thus the UAVs can be utilized for performing data harvesting and dissemination. For a complete data transmission process, the sensors of each IoT cluster will form a GVAA and select a single UAV from the UAV swarm as the receiver to transmit the sensed data. Afterward, the UAVs in the swarm should first broadcast the received data to other UAVs, and these UAVs will form different AVAAs to disseminate data to different remote BSs sequentially.

\par Without loss of generality, we consider a 3D Cartesian coordinate system, and the positions of the $i$th sensor of the $h$th IoT cluster, the $k$th BS and the eavesdropper $E$ are represented as $(x^{SN}_{h, i}, y^{SN}_{h, i}, 0)$, $(x^{B}_{k}, y^{B}_{k}, 0)$, and $(x^{E}, y^{E}, 0)$, respectively. Subsequently, we detail the key models with respect to signaling and transmission, including the virtual antenna array, transmission, and aerial broadcast models.

%
%
\subsubsection{Virtual Antenna Array Models}
\label{sssec:AF}

\par In a virtual antenna array model, each element with a single antenna (e.g., UAV or sensor) is an array element. After being carefully designed, the virtual antenna array will generate the desired beam pattern. In the following, we first introduce the GVAA and AVAA models, and then show the antenna gain model of virtual antenna arrays.

\par \textit{(i) GVAA Model:} We use the array factor to evaluate the signal strength of GVAAs in different directions. First, we let $(\theta, \phi)$ denote any direction centered on any GVAA, where $\theta \in [0, \pi]$ and $\phi \in [-\pi, \pi]$ are the elevation and azimuth angles, respectively. Moreover, we use excitation current weight to reflect the transmission power of a distributed antenna of sensors or UAVs. We denote the excitation current weight of $i$th sensor in $h$th cluster as $I_{i,h}^{SN}$. In addition, let a binary variable $D_{i,h}$ denote whether the $i$th sensor is selected to form the $h$th GVAA, and let $(x_{i, h}^{SN}, y_{i, h}^{SN}, z_{i, h}^{SN})$ be the 3D coordinates of the $i$th sensor in the $h$th GVAA. Then, the array factor of the GVAA is given by
\begin{equation}
   \begin{aligned}
	&A F_{h}^{SN}(\theta, \phi)= \\ &\sum_{i=1}^{N_{S N}} D_{i, h} I_{i, h}^{SN} e^{j^{u}\left[k_{c}\left(x_{i, h}^{SN} \sin \theta \cos \phi+y_{i, h}^{SN} \sin \theta \sin \phi+z_{i, h}^{SN} \cos \theta\right)+\delta_{i, h}\right]},
   \end{aligned}
   \label{eq:AF-GVAA}
\end{equation}

\noindent where $k_c=2\pi/\lambda$ is the phase constant, $\lambda$ is the wavelength, and $j^u$ is the imaginary unit. Since the sensors are fixed, the positions of sensors cannot be optimized. Thus, we can know that the array factors of GVAAs are determined by the selection of sensor and excitation current weights of selected sensors. Moreover, $\delta_{i, h}$ is the phase of the $i$th sensor in the $h$th cluster, which is given by
\begin{equation}
   \begin{aligned}
	&\delta_{i, h} = \\ &-\frac{2 \pi}{\lambda}\left(x_{i, h}^{SN} \sin \theta_{0} \cos \phi_{0}+y_{i, h}^{SN} \sin \theta_{0} \sin \phi_{0}+z_{i, h}^{SN} \cos \theta_{0}\right),
   \end{aligned}
   \label{eq:phase}
\end{equation}

\noindent where $(\theta_{0},\phi_{0})$ is the direction of the target receiver.

\par \textit{(ii) AVAA Model:} Similar to GVAA, we use the array factor to show signal distributions of AVAAs. Let $I_{j,k}^{UAV}$ and $P_{j, k}=(x_{j, k}^{U}, y_{j, k}^{U}, z_{j, k}^{U})$ denote the excitation current weight and position of the $j$th UAV for communicating with the $k$th BS, then the array factor of $k$th AVAA is given by
\begin{equation}
   \begin{aligned}
A&F_{k}^{U}(\theta, \phi)= \\ &\sum_{i=1}^{N_{U A V}} I_{j, k}^{UAV} e^{j^{u}\left[k_{c}\left(x_{j, k}^{U} \sin \theta \cos \phi+y_{j, k}^{U} \sin \theta \sin \phi+z_{j, k}^{U} \cos \theta\right)+\delta_{j, k}\right]},
    \end{aligned}
   \label{eq:AF-AVAA}
\end{equation}

\noindent where $\delta_{j, k}$ is the phase of $j$th UAV for communicating with $k$th BS. Different from GVAA, the UAVs have high mobility. Thus, we can derive that the array factors of AVAAs are determined by 3D coordinates and excitation current weights.

\par In both GVAA and AVAA, we assume that the perfect quantized version of the actual channel state information (CSI) is collected via the method in \cite{Ahmad2022CSI} to guide the wireless antenna transmission. Moreover, we also assume that the array elements of the virtual antenna arrays are synchronized in terms of the time and initial phase via the synchronization protocols and methods in~\cite{Mohanti2019} and~\cite{Boyle2017}. Note that we will evaluate the effect of imperfect CSI and synchronization in Section~\ref{sec:discussion}. Moreover, we consider that data sharing among sensors within the same cluster is accomplished by using the method in~\cite{Feng2010, Feng2013} (The main steps and overhead of this step will be discussed in Section~\ref{sec:discussion}).

\par \textit{(iii) Antenna Gain:} the antenna gain towards the receiver can be calculated by the array factor of the virtual antenna array~\cite{Mozaffari2019}, i.e.,
\begin{equation}
     \begin{aligned}
         G_{0} =
          &\frac{4\pi\left|AF(\theta_{0}, \phi_{0})\right|^2w{(\theta_{0}, \phi_{0})}^2}{\int_0^{2\pi}\int_0^\pi\left|AF(\theta, \phi)\right|^2w{(\theta, \phi)^2}\sin\theta\text{d}\theta\text{d}\phi}\eta,
     \end{aligned}
     \label{eq:Gain-BS}
\end{equation}
\noindent where ($\theta_{0}$, $\phi_{0}$) represents the direction towards the receiver. Moreover, $w{(\theta,\phi)}$ denotes the magnitude of the far-field beam pattern of each antenna element, and $\eta \in [0, 1]$ is the efficiency of the antenna array \cite{balanis2005antenna}.

%
%
\subsubsection{Transmission Model Based on CB}
\label{sssec:A2G_G2A}

\par The signals from the virtual antenna array are faded by the channel and decoded by the receiver. We adopt a probability LoS propagation model of 3GPP for both G2A and A2G transmission links since it can capture more practical effects. As suggested by 3GPP~\cite{TechnicalSpecificationGroup2017}, the LoS probability is mainly dependent on the altitude of the UAV $H_{U}$ and horizontal distance between the transmitter and receiver $d_{2\mathrm{D}}$, i.e.,
\begin{equation}
	\label{eq:probability_LoS}
	P_{\mathrm{LoS}}=\left\{\begin{array}{ll}
	P_{\mathrm{LoS}, \mathrm{ter}}, & 1.5 \mathrm{~m} \leq H_{U} \leq H_{1} \\
	P_{\mathrm{LoS}, \mathrm{U}}\left(d_{2\mathrm{D}}, H_{U}\right), & H_{1} \leq H_{U} \leq H_{2} \\
	1, & H_{2} \leq H_{U} \leq 300 \mathrm{~m}
	\end{array}\right.,
\end{equation}
\noindent where $P_{\mathrm{LoS}, \mathrm{ter}}$ is the LoS probability for conventional terrestrial networks. Moreover, $P_{\mathrm{LoS}, \mathrm{U}}\left(d_{2\mathrm{D}}, H_{U}\right)$ can be obtained as:
\begin{equation}
	\label{eq:los_p_2}
	\begin{aligned}
&P_{\mathrm{LoS}, \mathrm{U}} \left(d_{2\mathrm{D}}, H_{U}\right)\\
& =\left\{\begin{array}{ll}
1, & d_{2\mathrm{D}} \leq d_{1} \\
\frac{d_{1}}{d_{2\mathrm{D}}}+\exp \left(\frac{-d_{2\mathrm{D}}}{p_{1}}\right)\left(1-\frac{d_{1}}{d_{2\mathrm{D}}}\right), & d_{2\mathrm{D}}>d_{1}
\end{array}\right.,
\end{aligned}
\end{equation}
\noindent where $p_1$ and $d_1$ are given by logarithmic increasing functions of $H_{U}$ as specified in~\cite{TechnicalSpecificationGroup2017}.

\par Then, the Non-LoS (NLoS) probability is given by $P_{NLoS} = 1 - P_{LoS}$. In this case, the channel power gain is mainly determined by two attenuation factors for LoS and NLoS links which are $\mu_{\mathrm{LoS}}$ and $\mu_{\mathrm{NLoS}}$, respectively, and it can be expressed as $g_{c}=K_{0}^{-1} d_{rec}^{-\alpha}\left[P_{\text {LoS }} \mu_{\mathrm{LoS}}+P_{\mathrm{NLoS}} \mu_{\mathrm{NLoS}}\right]^{-1}$, where $K_{0}$ is the path-loss constant, and $d_{rec}$ is the distance between the transmitter and the receiver.

\par Thus, according to the antenna gain $G_{0}$ and channel gain $g_c$, the achievable rate from an antenna array to a receiver can be obtained as follows~\cite{Mozaffari2019, Sun2023}:
\begin{equation}
    \begin{aligned}
       R=B\log_2\left(1+\frac{P_{CB} G_{0}}{g_c\sigma^2}\right),
    \end{aligned}
    \label{eq:Transmission-rate-BS}
\end{equation}
\noindent where $B$ denotes bandwidth. Moreover, $P_{CB}=\sum I_i^2 P_{max} $ is the total transmit power of the antenna array, in which $I_i\sqrt{P_{max}}$ denotes excitation current of the $i$th element, where $I_i$ and $P_{max}$ are excitation current weight and maximum transmit power of the $i$th element, respectively. Note that we denote the transmission rates from the IoT cluster $h$ to the aerial UAV and from the AVAA to the $j$th BS as $R^{G2A}_h$ and $R^{A2G}_j$, respectively.

%
%
\subsubsection{Aerial Broadcast Model}
\label{sssec:broadcast}

\par After receiving the full data from GVAAs, the UAV receivers will broadcast the data to all UAVs. Due to the high altitude of UAVs, the aerial transmission should follow an LoS channel condition~\cite{Rieth2014A2A}. Let $\{A_1, A_2, ..., A_{N_{IoT}} | h \in \mathcal{V}, A_h \in \mathcal{U}\}$ denote the UAV set that receives data from GVAA and will broadcast the data to all UAVs (see Table~\ref{table:notations}). As for a UAV $A_h \in \mathbb{A}$, we use $d_{h,j}^{A2A}$ to denote the distance between the UAV receiver $A_h$ and the $j$th UAV of the swarm, then the transmission rate is given by
\begin{equation}
    \begin{aligned}
	R_{h, j}^{A2A}=B \log _{2}\left(1+\frac{P_{h} K_{0} d_{h, j}^{A2A^{-\alpha}}}{\sigma^{2}}\right),
    \end{aligned}
    \label{eq:broadcast-rate-BS}
\end{equation}

\noindent where $P_h$ is the transmission power of UAV $A_h$. Due to the broadcast nature, the actual broadcast rate should be the slowest rate $R_{\min h}^{A2A}$ obtained by all the UAVs, so that these UAVs can receive data simultaneously. From this model, we derive that the broadcast rate is determined by the transmission distance, which is related to the positions of the UAV receivers. \footnote{Note that there may be other advanced methods that can complete the data broadcasting, e.g., assigned orthogonal channels to UAVs for achieving interference-free~\cite{Mozaffari2019COMST}. However, like our work, the performance of these methods is also depended on the transmission distance between the source and receivers. Thus, some novel and advanced data broadcasting protocols can be embedded in our model directly, which means that our framework has some expandability.}

%
%
\subsection{Energy Cost Model of UAVs for CB}
\label{ssec:energy_consumption}

\par In this paper, we employ the typical rotary-wing UAVs since they are with high maneuverability and able to hover. In this work, we omit the communication energy cost of UAVs, as it is several orders of magnitude lower than that of propulsion~\cite{Zeng2019}. The propulsion energy consumption of a rotary-wing UAV in the two-dimensional (2D) horizontal plane is mainly determined by the speed $v$, which can be expressed as follows~\cite{Zeng2019}:
\begin{equation}
  \label{eq:energy-1}
   \begin{aligned}
      P_{UAV}\left(v\right)=&P_{B}\left(1+\frac{3v^2}{v_{tip}^2}\right)+P_{I}\left(\sqrt{1+\frac{v^4}{4v_0^4}}-\frac{v^2}{2v_0^2}\right)^{1/2}+\\&\frac{1}{2}d_0\rho sAv^3,
   \end{aligned}
\end{equation}
\noindent Other parameters of Eq. \eqref{eq:energy-1} are related to the UAV type and environment. Specifically, $P_{B}$ and $P_{I}$ are two constants that indicate the blade profile and induced powers, respectively, and $v_{tip}$, $v_{0}$, $d_{0}$, $s$, $\rho$, and $A$ are the parameters that represent tip speed of the rotor blade, mean rotor induced velocity in hovering, fuselage drag ratio, rotor solidity, air density, and rotor disc area, respectively.

\par Moreover, according to~\cite{Zeng2019a}, the propulsion energy consumption of 3D UAV trajectory can be extended from Eq. \eqref{eq:energy-1} by considering kinetic energy and potential energy theorems, i.e.,
\begin{equation}
  \label{eq:energy-2}
   \begin{aligned}
     E_{UAV}(T) \approx &\int_0^TP( v(t))dt+{\frac12m_{UAV}(v{(T)}^2- v{(0)}^2)}+\\
     &{m_{UAV}g(h(T)-h(0))},
   \end{aligned}
\end{equation}
\noindent where $v(t)$ is the instantaneous UAV speed of time $t$. Moreover, $T$, $m_{UAV}$, and $g$ are the end time of the flight, aircraft mass of the UAV, and gravitational acceleration, respectively.

\par As can be seen, the energy cost of a UAV is determined by its position change and speed control. As for UAV-enabled CB, the energy cost of the UAV swarm can be divided into two parts, i.e., hovering transmission energy $E^{tran}$ and propulsion energy for performing virtual antenna array $E^{perf}$. The former is determined by the lasting time of the transmission, and the latter is correlated to the speed control scheme when the trajectory is fixed. In this work, we follow the speed control scheme mentioned in~\cite{Sun2021}, and thus we can obtain the optimal speeds of UAVs for constructing a virtual antenna array according to the virtual antenna array performing time by using the method in~\cite{Sun2021}. Correspondingly, we denote the hovering transmission energy of the $j$th UAV for communicating with $k$th BS as $E^{tran}_{j,k}$, and represent the propulsion energy for performing virtual antenna array of the $j$th UAV for communicating with $k$th BS as $E^{perf}_{j,k}$.

%
%
\section{Problem Formulation and Analysis} 
\label{sec:problem_formulation_and_analysis}

\par The main goal of the considered system is to minimize the transmission time and energy costs of sensors and UAVs for accomplishing the data harvesting and dissemination tasks. On the other hand, the detected eavesdropper $E$ cannot be ignored for considering the secure performance of the communication. These goals can be achieved by jointly improving the performances of the sensor-enabled GVAAs and UAV-enabled AVAAs. Specifically, the beam patterns of the GVAAs and AVAAs can be jointly optimized to obtain higher directivity signals towards the targeted receivers and suppress the directivity towards the detected eavesdropper, thereby enhancing the transmission rates of receivers while decreasing the signal strength towards the eavesdropper.

\par To this end, a part of suitable sensors need to be selected and their optimal excitation current weights should be determined to achieve a better beam pattern for improving the performance of sensor-enabled GVAA of each IoT cluster. For the UAV-enabled AVAAs, the UAVs can move to better positions and use the optimal excitation weights to construct the AVAAs for CB. However, the UAV position changes will inevitably result in an increase of energy cost, thereby reducing the network service time. Thus, the above objectives should be comprehensively considered to find a reasonable trade-off between them.

\par Moreover, each sensor-enabled GVAA of different IoT clusters needs to select an appropriate UAV receiver. Therefore, as for the G2A transmission, selecting the suitable UAV (e.g., with reasonable link distance and angle) as the aerial receiver can affect the communication performance directly, and thus it needs to be carefully designed. In addition, in the A2G transmission, the data should be disseminated to multiple remote BSs, and the mainlobe of the UAV-enabled AVAA can point to only one BS at each time, which means that the UAVs need to re-position themselves over time to cater to different BSs. Thus, the order of communication with these remote BSs should be also properly designed, as this can affect the energy cost of UAVs.

\subsection{Definitions}
\label{sssec:notations_and_definitions}

\par In this subsection, we first present several essential definitions that related to the aforementioned goals. Table \ref{table:notations} contains a summary of decision variables used below, and these decision variables are further explained in Fig. \ref{fig:hybrid-solution} for a more intuitive expression.

\begin{table*}
    \centering
    \caption{Notations of the decision variables}
    \label{table:notations}
    \begin{tabularx}{\textwidth}{p{1cm}p{3cm}XX}
        \toprule
        \textbf{Variable set} &  \textbf{Variable elements} & \textbf{Physical meanings } & \textbf{Instance}  \\
        \midrule
        $\mathbb{D}$
        & $\{ D_{i,h} | \forall i \in \mathcal{W}_h, \forall h \in \mathcal{V}\}$
        & $\mathbb{D}$ represents which sensors are selected, while $D_{i,h}$ denotes the $i$th sensor of the $h$th sensor is whether selected.
        & Akin to Fig. \ref{fig:hybrid-solution}, if $D_{1,3}=1$, then it indicates the first sensor of the third sensor is selected for CB.
        \\
        \midrule
        $\mathbb{I^{SN}}$
        & $\{ I_{i,h}^{SN} | \forall i \in \mathcal{W}_h, \forall h \in \mathcal{V}\}$
        & $\mathbb{I^{SN}}$ represents the excitation current weights of the sensors, while $I_{i,h}^{SN}$ denotes the excitation current weight of the $i$th sensor in the $h$th sensor.
        & As shown in Fig. \ref{fig:hybrid-solution}, if $I_{1,3}^{SN}=0.9$, then it means the excitation current weight of the first sensor in the third sensor is $0.9$.
        \\
        \midrule
        $\mathbb{P}$
        & $\{ P_{j,k} | \forall j \in \mathcal{U}, \forall k \in \mathcal{B} \}$
        & $\mathbb{P}$ represents the positions of the UAVs for serving the BSs, while $P_{j,k}$ denotes the position of the $j$th UAV for communicating the $k$th BS.
        & Similar to Fig. \ref{fig:hybrid-solution}, if $P_{1,3}=(20,20,110)$, then it indicates the 3D positions of the first UAV for communicating the third BS is $(20,20,110)$.
        \\
        \midrule
        $\mathbb{I^{UAV}}$
        & $\{ I_{j,k}^{UAV} | \forall j \in \mathcal{U}, \forall k \in \mathcal{B} \}$
        & $\mathbb{I^{UAV}}$ represents the excitation current weights of the sensors, while $I_{j,k}^{UAV}$ denotes the excitation current weight of the $j$th UAV for serving the $k$th BS.
        & Akin to Fig. \ref{fig:hybrid-solution}, if $I_{1,3}^{UAV}=0.9$, then it indicates the excitation current weight of the $j$th UAV for serving the $k$th BS is $0.9$.
        \\
        \midrule
        $\mathbb{A}$
        & $\{A_1, A_2, ..., A_{N_{IoT}} | h \in \mathcal{V}, A_h \in \mathcal{U}\}$
        & $\mathbb{A}$ represents the aerial reviver UAVs of the IoT clusters, while $A_h$ denotes the aerial reviver UAV of the $h$th IoT.
        & As shown in Fig. \ref{fig:hybrid-solution}, if $\mathbb{A}= \{3, 2 \}$, then it means the first and second IoT clusters select the UAV $3$ and $2$ as the aerial reviver UAVs, respectively.
        \\
        \midrule
        $\mathbb{Q}$
        & $\{Q_1, Q_2, ..., Q_{N_{BS}} | k \in \mathcal{B}, Q_k \in \mathcal{B}\}$
        & $\mathbb{Q}$ represents the order that the UAV-enabled AVAAs disseminate data to the different terrestrial BSs, while $Q_k$ denotes the $k$th serving $Q_k$ BS.
        & Akin to Fig. \ref{fig:hybrid-solution}, if $\mathbb{Q}= \{3, 2, ..., 5 \}$, then it means the third BS is served in first order, the second BS is served in the second order, and so on.
        \\
        \bottomrule
    \end{tabularx}%
    \label{tab:addlabel}%
\end{table*}%

%
%
\vspace{+1 mm}
\begin{definition}[Mission Completion Time]
	We define mission completion time $T^{MCT}$ as the time from the sensor-enabled GVAAs start the transmission until the last BS receives the entire data.
\end{definition}
\vspace{+1 mm}

\par The mission consists of three components, i.e., \emph{\textbf{(a)}} G2A transmission process, \emph{\textbf{(b)}} air to air (A2A) transmission process, and \emph{\textbf{(c)}} A2G transmission process. Thus, $T^{MCT}$ contains the G2A transmission time $T^{G2A}$, A2A transmission time $T^{A2A}$ and A2G transmission time $T^{A2G}$, respectively, i.e., $T^{MCT}= T^{G2A}+T^{A2A}+T^{A2G}$. Then, we define the G2A transmission time, A2A transmission time and A2G transmission time as follows.
\vspace{+1 mm}

%
%
\vspace{+1 mm}
\begin{definition}[G2A Transmission Time]
	G2A transmission time $T^{G2A}$ denotes the time consumed by the sensor-enabled GVAAs for transmitting data to the UAV receivers.
\end{definition}
\vspace{+1 mm}

\par Let $N_{\text {data }_{h}}$ and $R^{G2A}_h$ denote the data that need to be transmitted and the G2A transmission rate of the $h$th IoT, respectively. Then, $T^{G2A}= \{ \widetilde{\text{Max}}(N_{\text {data }_{h}} / R^{G2A}_h| h \in \mathcal{V} )$, wherein $\widetilde{\text{Max}}()$ is maximizing operator for calculating the maximum value of the elements in a vector.

\par $T^{G2A}$ can be calculated by using Eqs. \eqref{eq:AF-GVAA}-\eqref{eq:Transmission-rate-BS} and the decision variables are derived as follows. Specifically, we only need to know $R^{G2A}$ of each IoT cluster since other parameters can be seen as constants. First, we derive the decision variables of the antenna gain of the GVAAs. As shown in Eqs. \eqref{eq:AF-GVAA} and \eqref{eq:Gain-BS}, the positions and excitation current weights of array elements are required. Note that not all sensors will participate in the CB process, which means that we need to select suitable sensors from all sensors for CB in a cluster. In this case, we can obtain the positions of array elements of this GVAA according to the sensor selection case $\mathbb{D}$, i.e., $\mathbb{D}$ is the decision variable (See Table \ref{table:notations} and Fig. \ref{fig:hybrid-solution} for details). Likewise, let the excitation current weights of sensors be $\mathbb{I^{SN}}$ which is another decision variable. Then, we also derive the decision variables of the channel fading process. Clearly, the key factor of this process is the positions of the aerial UAV receivers. Let the positions of UAVs denoted as $\mathbb{P}$ be the decision variable, the key factor can be converted to which UAV is selected as the aerial receiver to the IoT cluster. Accordingly, we denote the UAV receiver selection case as $\mathbb{A}$, and $\mathbb{A}$ is the decision variable.
\vspace{+1 mm}

%
%
\vspace{+1 mm}
\begin{definition}[A2A Transmission Time]
	A2A transmission time $T^{A2A}$ is the time consumed by the UAVs for broadcasting data to all UAVs in the AVAA, i.e., the time of data fusion in the UAV swarm for CB.
\end{definition}
\vspace{+1 mm}

\par In this process, the aerial receivers will broadcast data to other UAVs. Let $R_{\min_h}^{A2A}$ be the minimum transmission rate obtained by these UAVs when the $h$th aerial receiver broadcasts the data. Then, $T^{A2A}=\sum_h^{N_{IoT}} {N_{\text {data }_{h}}}/{R_{\min _{h}}^{A2A}}$. We can derive that $T^{A2A}$ can be determined by the UAV receiver selection case $\mathbb{A}$ and the position of UAVs $\mathbb{P}$. The reason is that aerial transmission is mainly related to the positions of the transmitter and receiver, and other parameters can be seen as constants.

%
%
\vspace{+1 mm}
\begin{definition}[A2G Transmission Time]
	A2G transmission time $T^{A2G}$ is the time consumed by the UAV-enabled AVAA for forwarding data to all the remote BSs.
\end{definition}
\vspace{+1 mm}

\par In this process, the UAVs firstly fly to the assigned position for constructing AVAAs and then hover for transmission. Thus, $T^{A2G}$ contains two components that are the AVAA performing time $T^{A2G}_{perf}$ and data transmission time between the AVAAs and all BSs $T^{A2G}_{tran}$.

\par Let $\mathbb{T}_{perf} = \{ T^{perf}_1, T^{perf}_2, ..., T^{perf}_{N_{BS}}\}$, where $T^{perf}_k$ denotes the time that the UAVs fly to the assigned positions for performing the AVAA so that communicating with the $k$th BS. Clearly, $\mathbb{T}_{perf}$ is the decision variable of $T^{A2G}$. On the other hand, according to~\cite{Sun2021}, we can set suitable $T^{perf}_k$ to optimally control the speeds of UAVs.

\par Moreover, $T^{A2G}_{tran}$ can be calculated by using Eqs. \eqref{eq:AF-AVAA}-\eqref{eq:Transmission-rate-BS} and the decision variables are derived as follows. Specifically, $T^{A2G}_{tran}= \widetilde{\text{Sum}}(\mathbb{T}_{tran})$, in which $\mathbb{T}_{tran} = \{ T^{tran}_1, T^{tran}_2, ..., T^{tran}_{N_{BS}}\}$, where $T^{tran}_k= (\sum_{h=1}^{N_{IoT}} N_{\text {data}_h})/R^{A2G}_k$, wherein $R^{A2G}_k$ is A2G transmission rate to the $k$th BS. We only need to derive $R^{G2A}$ of each AVAA since other parameters can be seen as constants. As shown in Eqs. \eqref{eq:AF-AVAA} and \eqref{eq:Gain-BS}, the positions of UAVs ($\mathbb{P}$) and excitation current weights of UAVs ($\mathbb{I^{UAV}}$) are decision variables. Moreover, the AVAA needs to transmit data to multiple BSs, whereas the mainlobe of AVAA can only point at a direction. Thus, the order of communication with these remote BSs denoted as $\mathbb{Q}$ is also the decision variable.

\subsection{Optimization Objectives}
\label{sssec:optimization_objective}

\par Based on the aforementioned definitions and analysis, the optimization objectives are presented as follows.

%
%
\par \emph{\textbf{Optimization Objective 1:}} The primary objective of this work is to complete the task as soon as possible thereby saving spectrum resources of terrestrial BSs, sensors, and UAVs. Thus, the first objective is to minimize the mission completion time, which can be expressed as follows:
\begin{equation}
   \begin{aligned}
   f_{1}(\mathbb{A}, \mathbb{D}, \mathbb{I^{SN}}, \mathbb{P}, \mathbb{I^{UAV}}, \mathbb{T}_{perf}, \mathbb{Q}&)=
   T^{MCT},
   \end{aligned}
	\label{eq: objective1}
\end{equation}
\noindent where $X= [\mathbb{A}, \mathbb{D}, \mathbb{I^{SN}}, \mathbb{P}, \mathbb{I^{UAV}}, \mathbb{T}_{perf}, \mathbb{Q}]$ is the full solution of the optimization problem.

%
%
\par \emph{\textbf{Optimization Objective 2:}} The SLL is a normalized description of the signal strength, and it is proportional to the signal strength in a certain direction. It can be seen from the Eqs. \eqref{eq:AF-GVAA}, \eqref{eq:AF-AVAA}, \eqref{eq:Gain-BS}, and \eqref{eq:Transmission-rate-BS} that suppressing SLL towards a direction can reduce the corresponding transmission rate to some extent. To ensure security performance, the SLLs towards the eavesdropper of sensor-enabled and UAV-enabled virtual antenna arrays should be minimized, which can be expressed as follows:
\begin{equation}
	\begin{aligned}
		f_{2}(&\mathbb{A}, \mathbb{D}, \mathbb{I^{SN}}, \mathbb{P}, \mathbb{I^{UAV}}, \mathbb{Q})=\\& \underbrace{ \sum_{k=1}^{N_{BS}}\frac{\left | AF^{U}_k\left ( \theta_{E_k}, \phi_{E_k} \right ) \right |}{AF^{U}_k\left ( \theta_{ML_{k}}, \phi_{ML_{k}} \right)}}_{\text{SLLs of AVAAs}} + \underbrace{ \sum_{h=1}^{N_{IoT}}\frac{\left | AF^{SN}_h\left ( \theta_{E_h}, \phi_{E_h} \right ) \right |}{AF^{SN}_h\left ( \theta_{ML_h}, \phi_{ML_h} \right)}}_{\text{SLLs of GVAAs}},
	\end{aligned}
	\label{eq: objective2}
\end{equation}
\noindent where $( \theta_{E_k}, \phi_{E_k} )$ and $(\theta_{ML_{k}}, \phi_{ML_{k}})$ are the directions of the eavesdropper and mainlobe of the UAV-enabled AVAA for communicating with the $k$th BS, respectively. Moreover, $( \theta_{ML_h}, \phi_{ML_h} )$ and $( \theta_{E_h}, \phi_{E_h} )$ are the directions of the mainlobe and eavesdropper of the $h$th sensor-enabled GVAA, respectively.

%
%
\par \emph{\textbf{Optimization Objective 3:}} To achieve efficient data harvesting and dissemination, the UAVs need to continuously change their positions to form different AVAAs. In this case, the UAVs will consume a certain amount of propulsion energy. Moreover, the UAV-enabled AVAAs need to hover to broadcast data and communicate with the remote BSs, which also are energy-consuming tasks. To save the energy cost of UAVs and thus extend the network lifetime, the third objective function is designed as follows:
\begin{equation}
\begin{aligned}
f_{3}(\mathbb{A}, \mathbb{P}, \mathbb{T}_{perf}, \mathbb{Q})= & \sum_{j=1}^{N_{U A V}} \sum_{k=1}^{N_{B S}} \underbrace{ E_{j, k}^{tran}(T_{k}^{tran})}_{\text{A2G transmission}}+\underbrace{E_{j, k}^{perf }(T_{k}^{perf})}_{\text{Motion for CB}} \\
& + \sum_{j=1}^{N_{UAV}} \underbrace{E_{j}^{A2A}(T^{A2A})}_{\text{A2A broadcasting}}.
\end{aligned}
\label{eq: objective3}
\end{equation}	

\noindent where $E_j^{A2A} (T^{A2A})$ is the energy cost of $j$th UAV for hovering $T^{A2A}$ time to broadcast or receive data. 

\par As can be seen, the three objectives are controlled by the mostly same decision variables. Moreover, they conflict with each other. Specifically, if the total SLLs of the eavesdropper ($f_2$) are suppressed, more transmit energy will be contained in the mainlobe, resulting in low antenna directivity. In this case, the transmission rates of the GVAA and AVAA will degrade, which means that the mission completion time ($f_1$) will increase. Likewise, suppose we reduce the mission completion time ($f_1$) via increasing UAV speed to reduce the motion time of the UAV, as shown in~\cite{Zeng2019}. In that case, the energy costs of the UAVs ($f_3$) will undoubtedly increase. In other words, we cannot find a solution to make $f_1$, $f_2$, and $f_3$ tend to be optimal simultaneously.

\par Accordingly, instead of using the weighted sum method, we formulate an MOP to find different trade-offs between the three optimization objectives as follows:
\begin{subequations}
  \label{eq:MOP-formulation}
  \begin{align}
    {\underset{X}{\text{min}}} \quad  & F=\{f_{1}, f_{2}, f_{3} \},\\
    \text{s.t.} \quad
	& D_{h, i} \in \{0, 1\}, \forall h \in \mathcal{V}, \forall i \in \mathcal{W}_h \label{eq:const1},\\
	& 0 \leqslant  I^{SN}_{h, i} \leqslant  1, \forall h \in \mathcal{V}, \forall i \in \mathcal{W}_h \label{eq:const2},\\
	& 0 \leqslant  I_{j, k}^{UAV} \leqslant  1, \forall j \in \mathcal{B}, \forall k \in \mathcal{U} \label{eq:const3},\\
	&  (x_{j, k}^{U}, y_{j, k}^{U}, z_{j, k}^{U}) \in \mathbb{R}^{3 \times 1}, \forall j \in \mathcal{B}, \forall k \in \mathcal{U}, \label{eq:const4}\\
	& A_h \in \mathcal{U}, \forall h \in \mathcal{V}, \label{eq:const5}\\
    & \mathbb{Q} \in \mathcal{Q}, \label{eq:const6}\\
    & d_{(j_1, j_2)} \geq d_{min},  \forall j_1, j_2 \in \mathcal{U} \label{eq:const7},
  \end{align}
\end{subequations}
\noindent where $\mathbb{R}^{3 \times 1}$ is the region that the UAVs can reach. Moreover, $\mathcal{Q}$ is the set of orders that the UAV-enabled AVAA communicates with $N_{BS}$ different BSs, which has $N_{BS}!$ possible permutations. In addition, the constraint \eqref{eq:const6} indicates that the minimum separation distance between two adjacent UAVs must be no less than $d_{min}$ to avoid collision. Furthermore, the formulated MOP is analyzed as follows.

\subsection{Problem Analysis}
\label{sssec:problem_analysis}

\par We further analyze the formulated MOP and provided some lemmas in this subsection as follows.

%
\begin{figure*}
  \centering
  \includegraphics[width=7in]{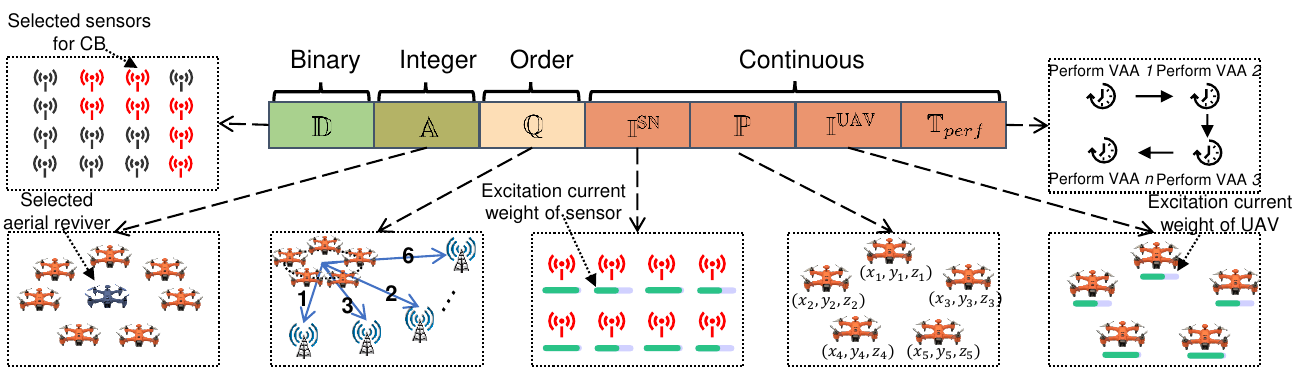}
  \caption{The solution structure of the formulated MOP. All decision variables are essential for controlling the key parameters of our considered system. }
  \label{fig:hybrid-solution}
\end{figure*}

\vspace{+1 mm}
\begin{lemma}
\label{lemma:NP-hard}
The optimization objectives of the formulated MOP shown in Eq. \eqref{eq:MOP-formulation} can be simplified as NP-complete problems. Accordingly, the formulated MOP shown in Eq. \eqref{eq:MOP-formulation} is NP-hard.
\end{lemma}
\vspace{+1 mm}
\begin{proof}
See Appendix A.1 of the supplemental material.
\end{proof}

\vspace{+1 mm}
\begin{lemma}
\label{lemma:large-scale}
The formulated MOP shown in Eq. (\ref{eq:MOP-formulation}) is a large-scale optimization problem, and its number of decision variables is $(1+2N_{SN})N_{IoT}+(4N_{UAV}+2)N_{BS}$.
\end{lemma}
\vspace{+1 mm}
\begin{proof}
See Appendix A.2 of the supplemental material.
\end{proof}
\vspace{+1 mm}

\begin{lemma}
The formulated MOP shown in Eq. (\ref{eq:MOP-formulation}) is a mixed-variable optimization problem with continuous, binary, integer, and order decision variables.
\label{lemma:hybrid-MOP}
\end{lemma}
\vspace{+1 mm}
\begin{proof}
See Appendix A.3 of the supplemental material.
\end{proof}
\vspace{+1 mm}

\par As shown in Lemma \ref{lemma:NP-hard}, the complexity of the formulated MOP increases significantly as the network scale grows, and thus it is difficult to find a deterministic algorithm to solve it. Moreover, Lemmas \ref{lemma:large-scale} and \ref{lemma:hybrid-MOP} demonstrates that some existing methods cannot solve the problem efficiently. Specifically, since our problem has massive decision variables and multiple objectives, reinforcement learning confronts issues of difficulty to converge and to determine objective weights. Likewise, as the formulated MOP is a mixed-variable optimization problem with continuous, binary, integer, and order decision variables in Lemma \ref{lemma:hybrid-MOP}, which can be simplified constraint programming problems~\cite{Meng2020}, it cannot be solved via convex methods directly or after relaxation.

\par In the following two sections, we aim to propose a swarm intelligence algorithm to control the decision variables of the formulated MOP. We first review the framework and some properties of the multi-objective optimization and swarm intelligence algorithm. Then, we enhance the algorithm to make it more suitable for solving our MOP.

%
%
\section{Preliminaries of Multi-objective Optimization and Swarm Intelligence} 
\label{sec:multi_objective_optimization_and_swarm_intelligence}

\par In this section, we present the preliminaries about multi-objective optimization and swarm intelligence algorithm to facilitate the understanding of our solving method. 

%
%
\subsection{Multi-objective Optimization}
\label{sub:multi_objective_optimization}

\par Multi-objective optimization is concerned with mathematical optimization problems involving more than one objective function to be optimized simultaneously. In multi-objective optimization, the comparison between different solutions can be achieved by \textbf{Pareto dominance} instead of arithmetic relational operators~\cite{Liu2022}. Let $\min F=[f_o(X) | o=1,2,...N_{obj} ]$ be an MOP with $N_{obj}$ objectives, Pareto dominance can be defined as follows:
\vspace{1mm}
\begin{definition}[Pareto dominance]
	A vector $X$ dominate $X'$ \\
	1. If $f_o(X) \le f_o(X')$ in all $N_{obj}$ optimization objectives, and\\
	2. There is at least one objective $o$ such that $f_o(X) < f_o(X')$.
\end{definition}
\vspace{1mm}

\par As such, the optimal solutions of an MOP are a set of solutions that do not Pareto dominate each other, which are with different trade-offs among the various optimization goals. Then, the decision-maker can flexibly select a suitable solution as the final solution according to the scenarios.

%
%
\subsection{Principles of Swarm Intelligence} 
\label{sub:principles_of_swarm_intelligence}

%
\begin{figure*}
  \centering
  \includegraphics[width=6.5in]{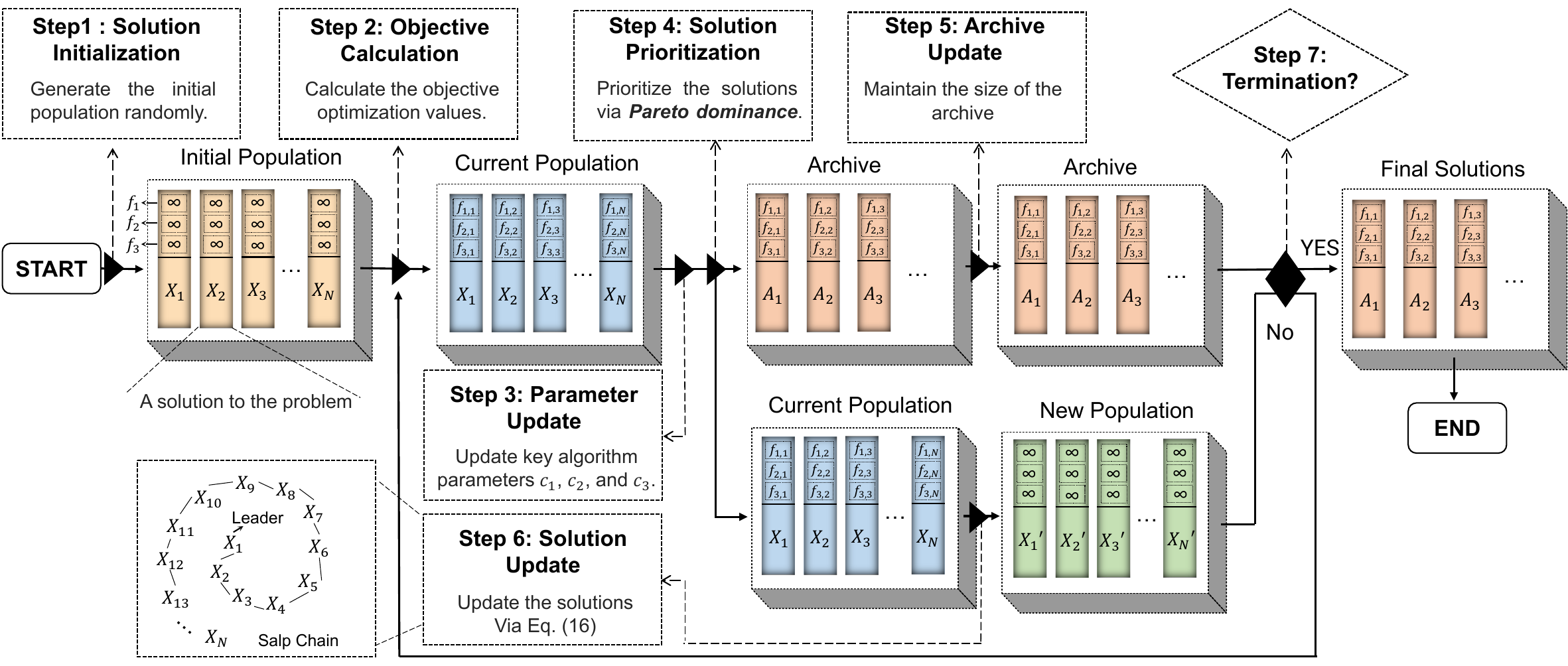}
  \caption{The algorithm framework of MSSA. Each cube is a population or archive, in which rectangles denote the candidate solutions with their three objective values.}
  \label{fig:algorithm-framework}
\end{figure*}

\par Swarm intelligence algorithms maintain and improve multiple candidate solutions iteratively, and they use heuristic swarm intelligence to guide the search. In this case, swarm intelligence algorithms have the potential to solve the formulated MOP since the algorithms can obtain candidate solutions in a limited time and cope with the NP-hardness. In solving MOP, multi-objective swarm intelligence can find solutions that dominate the others continuously. Among various swarm intelligence algorithms, the multi-objective salp swarm algorithm (MSSA), which is inspired by the swarm behaviors of salp when navigating and foraging in oceans, is demonstrated to outperform other algorithms in some applications~\cite{Abualigah2020}. As shown in Fig.~\ref{fig:algorithm-framework}, the main steps of MSSA are as follows.

\begin{itemize}
	\item \textit{Step 1 (Population Initialization):} MSSA initializes a population consists of candidate solutions via various random manners, and each solution (i.e., salp in MSSA) must be a potential solution to the MOP.

	\item \textit{Step 2 (Solution Evaluation):} The candidate solutions will be evaluated by calculating the objective values via Eqs. \eqref{eq:AF-GVAA}-\eqref{eq:MOP-formulation}.

	\item \textit{Step 3 (Parameter Update):} MSSA updates parameters $c1$, $c2$, and $c3$ which can determine the exploration and exploitation abilities, i.e.,
	\begin{equation}
	\label{eq:c1}
	c1=2e^{-\left(\frac{4 t}{t_{max}}\right)^{2}},
	\end{equation}
	\noindent where $t$ is the iteration number and $t_{max}$ is the maximum iteration. Moreover, $c2$ and $c3$ are two random numbers between 0 and 1 newly generated in each iteration.

	\item \textit{Step 4 (Solution Prioritization):} The algorithm will maintain a candidate solution set, namely, archive, to save the solutions that Pareto dominate the others. In this step, the candidate solutions and the solution in the archive will be sorted by Pareto dominance, and the better solutions will be maintained in the archive.

	\item \textit{Step 5 (Archive Update):} The MSSA will remove some candidate solutions from the archive via hypercubes mechanism~\cite{Mirjalili2017a} if the archive is oversize.

	\item \textit{Step 6 (Solution Update):} The current population is updated by using the inspired method of salp swarm. Specifically, the population will construct a salp chain and be divided into a leader and several followers, in which the leader guides the update of the followers. Mathematically, the update method of the $n$th dimension of the leader is given by
	\begin{equation}
	\begin{aligned}
	\label{eq:SSA-update-leader}
		&x_{1, n}=\\ &\left\{\begin{array}{cc}
	X_{best}(n)+c_{1}\left(\left(u b_{n}-l b_{n}\right) c_{2}+l b_{n}\right) & c_{3} \geq 0 \\
	X_{best}(n)-c_{1}\left(\left(u b_{n}-l b_{n}\right) c_{2}+l b_{n}\right) & c_{3}<0
	\end{array}\right.,
	\end{aligned}
	\end{equation}
	\noindent where $x_{1, n}$ is the $n$th dimension of the first solution (i.e., the leader), $ub_{n}$ and $lb_{n}$ are the upper and lower bounds of the $n$th dimension, respectively, and $X_{best}(n)$ is the $n$th dimension of the best solution of the population. As such, the $m$th followers of the population can be updated as $x_{m, n}=\frac{1}{2}\left(x_{m,n}+x_{m-1,n}\right)$.

	\item \textit{Step 7 (Termination):} Determine whether the termination condition is reached (e.g., reach the maximum iteration number or convergence). If no, repeat steps 2-7. If yes, the solutions within the archive are the final solutions, which are feasible engineering solutions with different trade-offs to the problem.

\end{itemize}

\par Like other swarm intelligence algorithms, MSSA only can handle one type of decision variable. Additionally, we review some general properties of swarm intelligence in solving various engineering problems.

\begin{itemize}

	\item \textit{Sensitive to the Initial Population:} The population states in different iterations can be modeled by a non-homogeneous Markov chain~\cite{He2001initialization}. The population in the $t$th iteration is affected by that of the $t-1$th iteration, which means that the subsequent populations are sensitive to the initial population. Thus, several works improve the algorithm performance by changing the initial population distribution~\cite{Kazimipour2014initialization}.

	\item \textit{Sensitive to Exploration and Exploitation Weights:} No-free lunch theory~\cite{Yang2014lunch} shows that all optimization algorithms perform equally when their performance is averaged across all possible problems. Thus, we should try to find better weights of exploitation and exploitation for a specific problem, as done in~\cite{Yang2014lunch}.

	\item \textit{Sensitive to Solution Structure:} The candidate solutions must have the same structure as the feasible solutions to the engineering optimization problem. Moreover, the hard-core constraints concerning the solution structure must be handled to keep candidate solutions available.

\end{itemize}

\par Next, we will introduce our enhanced algorithm for solving the formulated MOP.

%
%
\section{Our Proposed EMSSA Method}
\label{sec:EMSSA}

\par In this section, an EMSSA method is extended from the conventional MSSA by enhancing Steps 1, 3, and 6 of MSSA, and proposing a constraint handling method. 

%
%
\subsection{Enhancement for Step 1}
\label{sssec:enhance-solution_initialization}

\par In EMSSA, an individual of the initial population must be the feasible solution to the formulated MOP, which is in the form of $X= [\mathbb{D}, \mathbb{A}, \mathbb{Q}, \mathbb{I^{UAV}}, \mathbb{P}, \mathbb{I^{UAV}}, \mathbb{T}_{perf}]$. This is because that if an individual of the initial population is not a feasible solution of the formulated MOP, its objective values cannot be obtained via Eqs. \eqref{eq:AF-GVAA}-\eqref{eq:MOP-formulation}. For instance, if $\mathbb{Q}$ which is the order part of an initial individual does not satisfy the constraint \eqref{eq:const5}, then this part loses its physical meaning as an order, resulting in the objective function failure. Moreover, as aforementioned, swarm intelligence is sensitive to the initial population due to its structure and principle. Thus, we propose a novel solution initialization method for hybrid and large-scale decision variables as follows.

\par A suitable initial population should be a balance of random and uniform. However, the formulated MOP is a mixed-variable optimization problem (see Lemma \ref{fig:hybrid-solution}), and it has four types of decision variables in the formulated MOP, i.e., binary ($\mathbb{D}$), integer ($\mathbb{A}$), order ($\mathbb{Q}$) and regular continuous dimensions ($\mathbb{I^{SN}}, \mathbb{P}, \mathbb{I^{UAV}}, \mathbb{T}_{perf}$). Thus, it is challenging to generate these mixed decision variables with random and uniform via conventional swarm intelligence. To overcome this issue, we first generate $\mathbb{D}$, $\mathbb{A}$, and $\mathbb{Q}$ that satisfy constraints \eqref{eq:const1}, \eqref{eq:const5}, and \eqref{eq:const6} via several random manners, respectively, and then employ pathological function to map the continuous dimensions.

\par \emph{(i) Initialize $\mathbb{D}$}: The node selection integer dimensions $\mathbb{D}$ represent which sensors are selected for performing CB of $N_{IoT}$ IoT clusters, which can be further expressed as 
\begin{equation}\label{eq:d-eq}
    \begin{aligned}
    \mathbb{D} =
    \begin{bmatrix}
    & D_{1, 1} &  D_{2, 1} & \cdots & D_{N_{SN}, 1} &\\
    &D_{1, 2} &  D_{2, 2} & \cdots & D_{N_{SN}, 2} &\\
    & \undermat{\text {Sensor selection case of an IoT for CB} }{D_{1, N_{IoT}} &  D_{2, N_{IoT}} & \cdots & D_{N_{SN}, N_{IoT}} } &\\
    \end{bmatrix},\\
    \end{aligned}
\end{equation}

\vspace{+5 mm}
\noindent As can be seen, the $h$th column of $\mathbb{D}$ represents the sensor selection case of the $h$th cluster. Thus, assuming that $N_{sel}$ sensors will be selected in the IoT cluster $h$, and accordingly, the $h$th column of $\mathbb{D}$ denoted as $\mathbb{D}_h$ is composed of $N_{sel}$ dimensions with value 1 and $N_{SN}-N_{sel}$ dimensions with value 0. Therefore, $\mathbb{D}_h$ can be initialed as follows:
\begin{equation} 	
	\label{eq:d_intial}
	\mathbb{D}_h= \widetilde{\text{Rand}_{1,0}}( N_{sel}, N_{SN}-N_{sel}),
\end{equation}
\noindent where $\widetilde{\text{Rand}_{1,0}}( a_0 , b_0 )$ is an operator that randomly combines $a_0$ ones and $b_0$ zeros. Clearly, the generated dimensions meet the constraint \eqref{eq:const1} and preserve randomness which is vital to the initial performance.

\par \emph{(ii) Initialize $\mathbb{A}$}: The aerial receiver selection integer dimensions ($\mathbb{A}$) is integer and can be initial as follows:
\begin{equation}
	\label{eq:a-in}
	\mathbb{A}= \widetilde{\text{Randi}}(\mathcal{B}, N_{IoT}),
\end{equation}
\noindent where $\widetilde{\text{Randi}}(\mathcal{B}, a_0)$ returns a vector with $a_0$ elements containing pseudo-random integers drawn from the discrete uniform distribution in the $\mathcal{B}$. As such, we can randomly generate the dimensions that meet the constraint \eqref{eq:const5}.

\par \emph{(iii) Initialize $\mathbb{Q}$}: The BS communication order dimensions $\mathbb{Q}$ is initial as follows:
\begin{equation}
	\label{eq:q-in}
	\mathbb{Q}= \widetilde{\text{Randperm}}(N_{BS}),
\end{equation}
\noindent where $\widetilde{\text{Randperm}}(a_0)$ can return a vector containing a random permutation of the integers from 1 to $a_0$ without repeating elements. Thus, we can randomly obtain the dimensions which meet the nature of order dimension directly.

\par \emph{(iv) Initialize Continuous Dimensions}: These dimensions generated by a pseudo-random number generator in MSSA may cause low convergence when dealing with large-scale problem~\cite{Kazimipour2014initialization}. Thus, after various practice attempts, we adopt the Weierstrass function which is an example of a real-valued pathological function to obtain a more homogeneous initialized population. Specifically, this initialization method can be expressed as follows:
\begin{equation}
	\label{eq:Weierstrass-intial}
	x_{n}^c=lb_{n}^c+f_{W}(l) \times (ub_{n}^c-lb_{n}^c),
\end{equation}
\noindent where $x_{n}^c$ is the $n$th continuous dimension, and the total number of continuous dimensions is set as $N_{cd}$. Moreover, $f_{W}(l)$ is the element of the $l$th dimension of the discrete vector of the Weierstrass function, wherein the Weierstrass function is defined in~\cite{eichler1982on}. By using this method, more homogeneous initialized continuous dimensions are generated which can improve the algorithm stability.

\par In summary, the proposed solution initialization method is shown in Algorithm \ref{Algorithm:solution-initialization}.

%
%
\begin{algorithm}
  \caption{Solution Initialization of EMSSA} \label{Algorithm:solution-initialization}
  Define the parameters: $N_{IoT}$, $N_{sel}$, $N_{BS}$, $N_{cd}$, etc.;\\

  $\mathbb{D} \leftarrow \varnothing$, $\mathbb{A} \leftarrow \varnothing$, $\mathbb{Q} \leftarrow \varnothing$;\\

  $\{\mathbb{I^{SN}}, \mathbb{P}, \mathbb{I^{UAV}}, \mathbb{T}_{perf}\} \leftarrow \varnothing$;\\

  \For{$h=1$ to $N_{IoT}$}
  {
  	Generate $\mathbb{D}_h$ via Eq. \eqref{eq:d_intial};\\
  	$\mathbb{D}= \mathbb{D} \cup \left\{\mathbb{D}_h \right\}$
  }

  Generate $\mathbb{A}$ and $\mathbb{Q}$ via Eqs. \eqref{eq:a-in} and \eqref{eq:q-in}, respectively;\\

  \For{$m=1$ to $N_{cd}$}
  {
  	Generate $x_m^c$ via Eq. \eqref{eq:Weierstrass-intial};\\
  	$\{\mathbb{I^{SN}}, \mathbb{P}, \mathbb{I^{UAV}}, \mathbb{T}_{perf}\} = \{\mathbb{I^{SN}}, \mathbb{P}, \mathbb{I^{UAV}}, \mathbb{T}_{perf}\} \cup \left\{ x_m^c \right\}$
  }

  Return $X= [ \mathbb{A}, \mathbb{D}, \mathbb{I^{SN}}, \mathbb{P}, \mathbb{I^{UAV}}, \mathbb{T}_{perf}, \mathbb{Q}]$;
\end{algorithm}

%
%
\subsection{Enhancement for Step 3}
\label{sssec:enhance-algorithm-parameter}

\par In this subsection, we present the enhancement methods for the key parameters of the proposed EMSSA. In EMSSA, we can carefully tune $c2$ and $c3$ to balance the exploration and exploitation abilities of the algorithm, and the reasons are as follows. As shown in Eq. \eqref{eq:SSA-update-leader}, the individuals of the population are updated by $X_{best}$ and new generated dimension $\left(\left(u b_{n}-l b_{n}\right) c_{2}+l b_{n}\right)$. The parameter $c3$ can determine whether we obtain the difference or the sum between these two dimensions, i.e., $c3$ can change the search direction of the population. Moreover, $c2$ determines the values of the newly generated dimension. When $c2$ is larger, the individual tends to change their dimensions, i.e., increase the exploration ability of the algorithm. Conversely, the individual tends to retain the information of $X_{best}$, i.e., increase the exploitation ability of the algorithm. Thus, $c2$ and $c3$ jointly show the balance of the exploration and exploitation abilities, thereby determining the performance of the algorithm.

\par However, in conventional MSSA, $c2$ and $c3$ are two random values with randomness and blindness, which leads to an imbalance between exploration and exploitation abilities when dealing with large-scale optimization problems such as the formulated MOP. In this case, an effective method is to map the $c1$ and $c2$ into the chaos domain for obtaining non-linear, ergodic, and stochastic weights between the exploration and exploitation phases. After extensive attempts,, we choose two chaos maps, i.e., Sine and Gauss maps, to reset the $c2$ and $c3$ as follows:
\begin{equation}
	\begin{aligned}
	\label{eq:parameter-update}
	c_2^t&= \frac{l}{4} \sin(\pi c_2^{t-1}), \quad \,\, l = 4, \\
	c_3^t&=\left\{\begin{array}{ll}
1, \quad\quad\quad\quad\quad c_3^{t-1}=0 \\
\frac{1}{\bmod \left(c_3^{t-1}, 1\right)}, \text { otherwise }
\end{array}\right.,
	\end{aligned}
\end{equation}
\noindent where $c_2^t$ and $c_3^t$ are the values of $c2$ and $c3$ in the $t$th iteration, respectively. By using this method, we can balance the exploration and exploitation abilities of the algorithm in a non-linear, ergodic, and stochastic manner. Since non-linear, ergodic, and stochastic properties could help the algorithm to search thoroughly in the irregular solution space, acquire preliminary knowledge of the solution space, and escape from local optima, respectively, our enhancement can increase the probability of finding the best solution~\cite{Tavazoei2007}.

%
%
\subsection{Enhancement for Step 6}
\label{sssec:Handle_of_Hybrid_Solution_Space}

\par In this subsection, we will introduce the enhancement methods for the solution update phase of the proposed EMSSA. The main principle of swarm intelligence is to use the complex relationship between solutions continuously to improve the quality of the candidate solutions. Thus, when updating the hybrid decision variables of the formulated MOP, it is essential to make each candidate solution exchange information with the population. However, the solution update method of the conventional MSSA only can update the continuous dimension (e.g., $\mathbb{I^{SN}}, \mathbb{P}, \mathbb{I^{UAV}}, \mathbb{T}_{perf}$ of this paper). In this case, the key measure is to continue the logic of EMSSA for solution update, i.e., establishing information exchanges between the solution with $X_{best}$ in the population, meanwhile, ensuring that the updated solution remains unchanged in its original structure.

\par Accordingly, we use different schemes to update binary, integer, and order decision variables according to the same control parameters. Let $X(\mathbb{D}, \mathbb{A}, \mathbb{Q}, \mathbb{I^{UAV}}, \mathbb{P}, \mathbb{I^{UAV}}, \mathbb{T}_{perf})$ and $X_{best}(\mathbb{D}, \mathbb{A}, \mathbb{Q}, \mathbb{I^{UAV}}, \mathbb{P}, \mathbb{I^{UAV}}, \mathbb{T}_{perf})$ be the solution to be update and best solution, the solution is updated as follows.

\par \emph{(i) Update $\mathbb{D}$}: Node selection binary dimensions of a solution ($X(\mathbb{D})$) represent the results that which sensors are selected for performing CB. In this work, we let $X(\mathbb{D})$ be affected by the same part of the best solution of the population $X_{best}(\mathbb{D})$ by performing a mutation, which can be expressed as follows:
\begin{equation}
	\label{eq:node-selection-update}
	X(\mathbb{D})= \left\{\begin{array}{ll}
	\widetilde{\text{Mut}}(X_{best}(\mathbb{D})), & \text{rand}> c1 \\
	\widetilde{\text{Mut}}(X(\mathbb{D})), &  \text{otherwise} \\
	\end{array}\right.,
\end{equation}
\noindent where $\text{rand}$ and $c1$ are a random number between 0 and 1 and the threshold parameter shown in Eq. \eqref{eq:c1}, respectively. Moreover, $\widetilde{\text{Mut}}()$ is a mutation operator for the binary vectors, which can be expressed as $\widetilde{\text{Mut}}(\{x_1, ..., x_{r1}, ..., x_{r2}, ..., x_{N_{SN}}\})$ = $\{x_1, ..., x_{r2}, ..., x_{r1}, ..., x_{N_{SN}}\}$, wherein $N_{SN} \geq r2 > r1 >0$. Note that we introduce the control parameters of updating the continuous decision variables to determine the rate of information exchange between the binary decision variables and the population. This is because we aim to keep the search direction and step to binary and continuous solution spaces be same.

\par \emph{(ii) Update $\mathbb{A}$}: Likewise, we let the aerial receiver selection integer dimensions of a solution ($X(\mathbb{A})$) be guided by $X_{best}(\mathbb{A})$ and random disturbance, which is controlled by $c1$, i.e.,
\begin{equation}
	\label{eq:aerial-update}
	X(\mathbb{A})= \left\{\begin{array}{ll}
	X_{best}(\mathbb{A}) & \text{rand}>c1 \\
	\widetilde{\text{Randi}}(\mathcal{B}, N_{IoT}) &  \text{otherwise} \\
	\end{array}\right..
\end{equation}

\par \emph{(iii) Update $\mathbb{Q}$}: Due to the non-repeating permutation nature, BS communication order dimensions ($X(\mathbb{Q})$) is intractable to be handled. To update this part, we introduce the partially mapped crossover (PMX)~\cite{Deep2012} operator to make the solution cross with $X_{best}(\mathbb{Q})$, which can be expressed as 
\begin{equation}
	\label{eq:order-update}
	X(\mathbb{Q})= \left\{\begin{array}{ll}
	PMX(X_{best}(\mathbb{Q}), X(\mathbb{Q}) ) & \text{rand} > c1 \\
	X(\mathbb{Q}) &  \text{otherwise} \\
	\end{array}\right..
\end{equation}

\par Overall, the enhanced solution update method is shown in Algorithm \ref{Algorithm:solution-update}. By using this method, the different parts of the solution can be guided by the $X_{best}$ and be controlled by the same parameter. As such, the algorithm searches in binary, integer, order, and continuous solution spaces in the same direction and step.

%
%
\subsection{Handling of Constraints}
\label{sssec:Handle_of_Constraint}

\par If the constraints are not satisfied, the updated solution may be beyond its feasible domain, resulting in algorithm failures. Thus, we handle the constraints of the formulated MOP in this part. Specifically, the constraints \eqref{eq:const1} and \eqref{eq:const5} can be met appropriately by using the solution initialization and update methods mentioned in Sections \ref{sssec:enhance-solution_initialization} and \ref{sssec:Handle_of_Hybrid_Solution_Space}, respectively. Moreover, akin to other swarm intelligence methods, the constraints \eqref{eq:const2}, \eqref{eq:const3}, and \eqref{eq:const4} can be handled by solving the upper and lower bounds of the decision variables $\mathbb{I^{SN}}, \mathbb{P}, \mathbb{I^{UAV}}$, and $\mathbb{T}_{perf}$, which is expressed as follows:
\begin{equation}
	\label{eq:constraints-c}
	x_{n}^c=\max\{\min\{ x_{n}^c, ub_{n}^c \},lb_{n}^c\},
\end{equation}
\noindent where $x_{n}^c$ is the $n$th dimension of the continuous parts.

\par Moreover, the constraint \eqref{eq:const6} follows a hard-core process with minimal separation distance $d_{min}$ between any two UAVs to avoid collision. However, the manner of deleting a solution if it does not satisfy this constraint may result in a large amount of wasted computational resources. Thus, we propose a novel scheme to handle the constraint and meanwhile retain the original information. Specifically, we first detect the collision and then use $L\acute{e}vy$ to re-position the solution for avoiding collision.

\par Concretely, the collision is determined by the solution part $\mathbb{P}$, and it can be detailed as follow:
\begin{equation}\label{eq:p-eq}
    \begin{aligned}
    \mathbb{P} =
    \begin{bmatrix}
    & P_{1, 1} &  P_{2, 1} & \cdots & P_{N_{UAV}, 1} &\\

    &P_{1, 2} &  P_{2, 2} & \cdots & P_{N_{UAV}, 2} &\\

    & \undermat{\text {Positions of UAV swarm for communicating a BS} }{P_{1, N_{BS}} &  P_{2, N_{BS}} & \cdots & P_{N_{UAV}, N_{BS}} } &\\

    \end{bmatrix},\\
    \end{aligned}
\end{equation}
\vspace{+5mm}

\noindent As can be seen, the $k$th row of the Eq. \eqref{eq:p-eq} is the positions that the UAVs communicate with the $k$th BS. Accordingly, we firstly judge UAV $j$ whether existing the collision conditions for communicating the $k$th BS as follows:
\begin{equation}
	\aleph=\left\{\begin{array}{ll}
	\text{true}, & ||P_{j,k}-P_{j_{min},k} ||>d_{min} \\
	\text{false}, & \text{otherwise} \\
	\end{array}\right.,
\end{equation}
\noindent where $P_{j_{min},k}$ is the position of the UAV which is nearest with the UAV $j$ when communicating with the $k$th BS. Then, if the $\aleph$ is true, we use $L\acute{e} v y$ flight to avoid collision as follows:
\begin{equation}\label{eq:handle-collision}
	P_{j,k}=  P_{j,k}+\alpha \oplus L \acute{e} v y(\lambda),
\end{equation}
\noindent where $ L\acute{e}vy(\lambda)$ is a random step value which is taken from $L\acute{e} v y$ distribution~\cite{Kaidi2022}, in which $ L\acute{e}vy(\lambda) \sim u=t^{-\lambda}(1<\lambda<3)$. By using this method, the collision part of the solution can adopt the short-distance and occasional long-distance searching of $L\acute{e} v y$ alternately, such that avoiding the collision and enhancing the search ability of the algorithm. In summary, the constraint handle method can be detailed in Algorithm \ref{Algorithm:constraint-handle}.

%
%
\begin{algorithm}
  \caption{Solution Update of EMSSA}\label{Algorithm:solution-update}
  Define the parameters: $X$, $X_{best}$, etc.;\\
  Update the $X(\mathbb{I^{SN}}, \mathbb{P}, \mathbb{I^{UAV}}, \mathbb{T}_{perf})$ by using Eq. \eqref{eq:SSA-update-leader};\\
  Update the $X(\mathbb{D})$ and $X(\mathbb{A})$ by using Eqs. \eqref{eq:node-selection-update} and \eqref{eq:aerial-update}, respectively;\\
  Update the $X(\mathbb{Q})$ by using Eq. \eqref{eq:order-update};\\
  Return $X$;
\end{algorithm}

%
%
\begin{algorithm}
  \caption{Constraint Handle Method}\label{Algorithm:constraint-handle}
  Define the parameters: $X$, $N_{BS}$, $N_{UAV}$, etc.;\\

  Handle constraints \eqref{eq:const2}, \eqref{eq:const3} and \eqref{eq:const4} by using Eq. \eqref{eq:constraints-c};\\

  \For{$k=1$ to $N_{BS}$}
  {
  	\For{$k=1$ to $N_{BS}$}
  	{
  		Calculate $\aleph$;\\
  		\If {$\aleph$}
  		{
  		Handle collision of UAV $k$ via Eq. \eqref{eq:handle-collision};\\
  		}
  	}
  }
  Return $X$;
\end{algorithm}

%
%
\subsection{Main Step and Complexity of the Proposed Algorithm}
\label{ssec:complexity}

\par Overall, the main structure of EMSSA is shown in Algorithm \ref{Algorithm:EMSSA}, in which $P_t$, $A_{t}$ and $F_{t}$ represent the population, archive and objective value set of the $t$th iteration, respectively, and $N_{dim}$, $N_{pop}$ and $t_{max}$ are the decision variable number of a solution, population size and maximum iterations, respectively.

\begin{algorithm}[]
\caption{EMSSA}\label{Algorithm:EMSSA}

	\KwIn{$P_0$, $N_{pop}$, $t_{max}$, $A_{0}$, etc.}
	\KwOut{$A_{t_{max}}$.}

	$P_0 \leftarrow \varnothing$, $A_{0} \leftarrow \varnothing$, $F_{0} \leftarrow \varnothing$;\\
	$c_1 \leftarrow 0$, $c_{2} \leftarrow 0$, $c_{3} \leftarrow 0$;\\

	\For{$m=1$ to $N_{pop}$}
	{

        Generate the $m$th solution $X_m$ via \emph{\textbf{Algorithm \ref{Algorithm:solution-initialization}}};\\
		$P_{0} \leftarrow P_{0} \cup\left\{X_m\right\}$;\\

	}

    \For{$t=1$ to $t_{max}$}
    {

    	\For{$m=1$ to $N_{pop}$}
    	{
			Calculate the objective values of $m$th solution $ f_{m}=[ f_{1_m},f_{2_m},f_{3_m} ]$;\\
			$F_{t} \leftarrow F_{t} \cup\left\{f_m\right\}$;\\
		}

        Update $A_t$ according to $F_{t}$ via \textbf{Pareto dominance} and \textbf{hypercube} mechanisms~\cite{Mirjalili2017a};\\

        Update $c1$, $c2$ and $c3$ via Eqs. \eqref{eq:c1} and \eqref{eq:parameter-update};\\

        \For{$m=1$ to $N_{pop}$}
        {
          Update the $m$th solution $X_m$ via \emph{\textbf{Algorithm \ref{Algorithm:solution-update}}};\\
          Handle constraint of $X_m$ via \emph{\textbf{Algorithm \ref{Algorithm:constraint-handle}}};\\
        }

        $F_{t+1} \leftarrow \varnothing$;\\
    }
   Return $A_{t_{max}}$;
\end{algorithm}

\vspace{+1 mm}
\begin{proposition}
The complexity of the proposed EMSSA is $\mathcal{O}(N_o N_{pop}^2)$.
\end{proposition}

\vspace{1 mm}
\begin{proof}
See Appendix A.4 of the supplemental material.	
\end{proof}


%
%
\section{Deployment Strategies} 
\label{sec:discussion}

%
%
\begin{figure}
  \centering
  \includegraphics[width=3.5in]{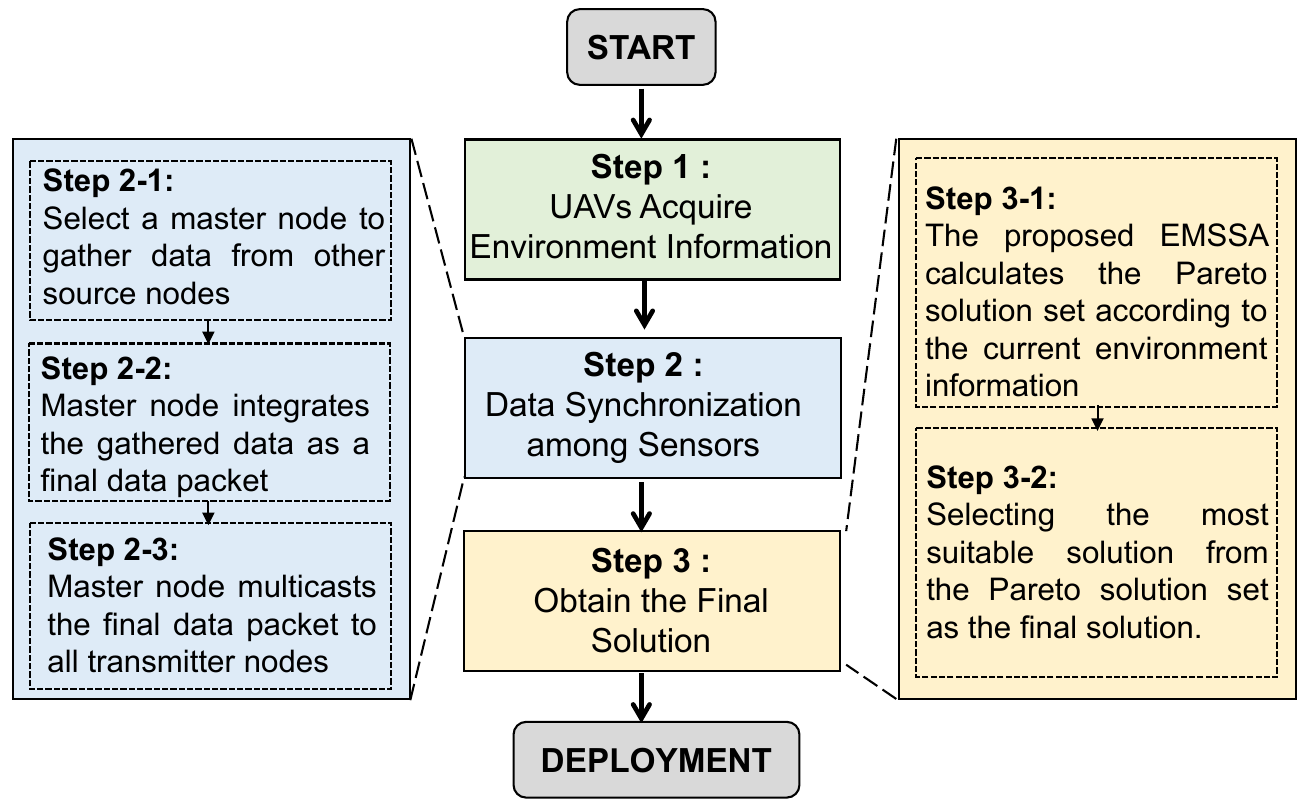}
  \caption{The main steps of our considered scheme for deploying CB-based data harvesting and dissemination system.}
  \label{fig:scheme}
\end{figure}

\par In this section, we discuss a feasible scheme when our method is used in practical applications. As shown in Fig.~\ref{fig:scheme}, this scheme consists of three main steps which are detailed as follows. 

\par \textit{1) UAVs Acquire Environment Information}. Most of the information, e.g., BS or sensor locations, is static and does not require active acquisition. However, since the eavesdropper's location may change, UAVs may need to detect the eavesdropper using cameras or radar. Note that using cameras or radar of UAVs requires relatively low energy compared to their propulsion energy~\cite{liu2022real,tarchi2017mini}. Moreover, our method does not require the radar and camera to work all the time. In our formulation, we mitigate eavesdropping by reducing the signal strength in the eavesdropper's direction (as shown in Eq.~\eqref{eq: objective2}). As a result, it is sufficient for UAVs to have the rough direction of the eavesdropper instead of precise positioning in real time using cameras and radar. Therefore, UAVs only need to intermittently use radar and cameras. In this case, the energy consumption of UAVs for using radar and cameras is rather small.

\par \textit{2) Data Synchronization among Sensors.} In this step, sensors within the same cluster perform data synchronization to have the same data by using the method in~\cite{Feng2010, Feng2013}. \textit{First}, a master node will be selected to gather data from other source nodes. \textit{Second}, the gathered data will be integrated as a final data packet. \textit{Finally}, the master node multicasts the final data packet to all transmitter nodes.

\par The overhead of this step is about 10-20 seconds which is little compared to the saved communication and motion time of UAVs by CB. Compared with other common schemes shown in simulations of Section 8.1.2 later, our method can achieve obvious time savings (ranging from 50\% to as high as 90\%). Moreover, the energy consumption associated with data sharing also is negligible which has been verified by reference~\cite{Feng2010}. 

\par \textit{3) Obtain the Final Solution.} In this step, the system obtains a final solution based on the proposed EMSSA and current environment information. \textit{First}, the proposed EMSSA shown in Algorithm~\ref{Algorithm:EMSSA} will find a set of candidate solutions (Pareto solution set, i.e., $A_{t_{max}}$) that represent the trade-offs between the three objectives. All the candidate solutions of the Pareto solution set are valuable and represent different trade-offs among the optimization objectives. \textit{Second}, selecting one solution from the Pareto solution set. Note that this aims to identify the most suitable trade-offs. If we care about an objective the most, we can select the solution with the best value for that specific objective. We can also use some low-complexity existing methods, such as gray relational analysis or additive weighting methods~\cite{Wang_2017,Ferreira2007}, to choose the final solution. Since the solutions in the Pareto solution set already exhibit balanced trade-offs, the selected solution will also possess favorable values for the remaining objectives.

%
%
\section{Simulation Results} 
\label{sec:simulation_results_and_analysis}

\par In this section, simulation results are provided to evaluate the effectiveness and superiority of the proposed EMSSA for solving the formulated MOP.

\par We consider two different scale scenarios, i.e., small-scale and large-scale scenarios, respectively. Specifically, in the small-scale scenario, the numbers of the IoT clusters in the monitor area, the sensors of each IoT cluster, the selected sensors of each IoT cluster, the UAVs, and the remote BSs are set as 2, 50, 10, 16, and 8, respectively, while in the large-scale, the corresponding numbers are set as 4, 50, 10, 32, and 8, respectively. Moreover, we also consider two UAV height settings for different communication channel conditions in both small-scale and large-scale scenarios, i.e., high altitude setting for an entire LoS channel and low altitude for a probability LoS channel, in which the UAV heights of different settings are set from 100 m to 120 m and from 70 m to 90 m, respectively. In addition, the monitoring area of each IoT cluster is set as 100 m $\times$ 100 m, and the UAVs are also distributed in a 100 m $\times$ 100 m area. The collision distance $d_{min}$ is set as 0.5 m, respectively. Furthermore, the carrier frequency $f_c$, bandwidth $B$, total transmit power of UAVs or sensors $P_i$, pathloss exponent $\alpha$, and total noisy power spectral density are 0.9 GHz, 2 MHz, $0.1$ W, 2, and -157 dBm/Hz, respectively. The parameters about propulsion energy consumption of UAV $v_{tip}$, $v_0$, $\rho$, $A$, $d_0$, and $s$ are set as 120 m/s, 4.03 m/s, 1.225 kg/$\text{m}^3$, 0.503 $\text{m}^3$, 0.6, and 0.05, respectively.

\par For comparison, several types of comparison algorithms and methods are introduced in this work. 

\begin{itemize}
	\item \textit{RandomLAA:} A benchmark method based on CB, i.e., selecting sensors to form a sensor-enabled GVAA randomly and making UAVs form a linear antenna array to perform the communication, namely, the RandomLAA method, is introduced. 

	\item \textit{Conventional MSSA:} We introduce the original version of MSSA as the comparison algorithm to show the effectiveness of our proposed enhancement measures.

	\item \textit{Newly Proposed or Classic Swarm Intelligence Algorithms:} We introduce some multi-objective swarm intelligence algorithms in existing works for comparisons, including conventional MSSA, multi-objective stochastic paint optimizer (MOSPO)~\cite{Khodadadi2022}, multi-objective multi-verse optimization (MOMVO)~\cite{Mirjalili2017}, multi-objective dragonfly algorithm (MODA)~\cite{Mirjalili2016}, and multi-objective particle swarm optimization (MOPSO)~\cite{Coello2002}, to solve the formulated MOP. These comparisons can show the performance of our proposed EMSSA.

	\item \textit{Non-CB Strategies:} Some benchmark strategies for data harvesting and dissemination, that are, the UAVs flying between sensors and BSs and the UAVs constructing a multi-hop flying ad-hoc network, are introduced for analyses.

\end{itemize}

\par Note that objective 1 shown in Eq.~\eqref{eq: objective1} is the most important objective in most scenarios. Thus, for all multi-objective optimization algorithms, we select the solution with the best objective 1 from the Pareto solution set as the final solution. 

%
%
\subsection{Performance Evaluation}
\label{ssec:Performance_evaluation}

\par In this section, we first show the comparison results between different CB-based methods. Then, we compare our proposed method with some other communication strategies. Finally, we verify the performance of our method in some special cases. Note that we also present some visualization results for an initial intuition, and they are provided in Appendix B of the supplemental material. 

%
%
\subsubsection{Comparison with Different CB-based Approaches}

%
\begin{figure*}
    \centering
	  \subfloat[Small-scale scenario under the entire LoS channel.]{
       \includegraphics[width=0.45\linewidth]{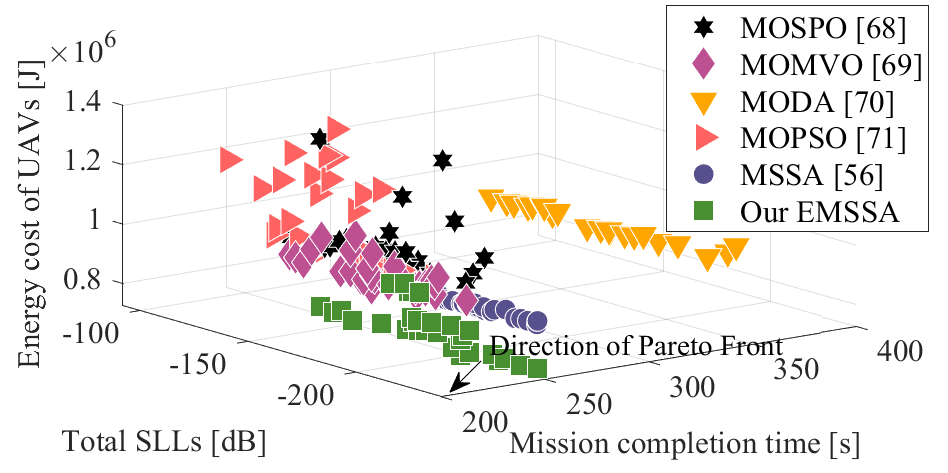}\label{fig:PS-results1}}
    \centering
	  \subfloat[Small-scale scenario under the probability LoS channel.]{
        \includegraphics[width=0.45\linewidth]{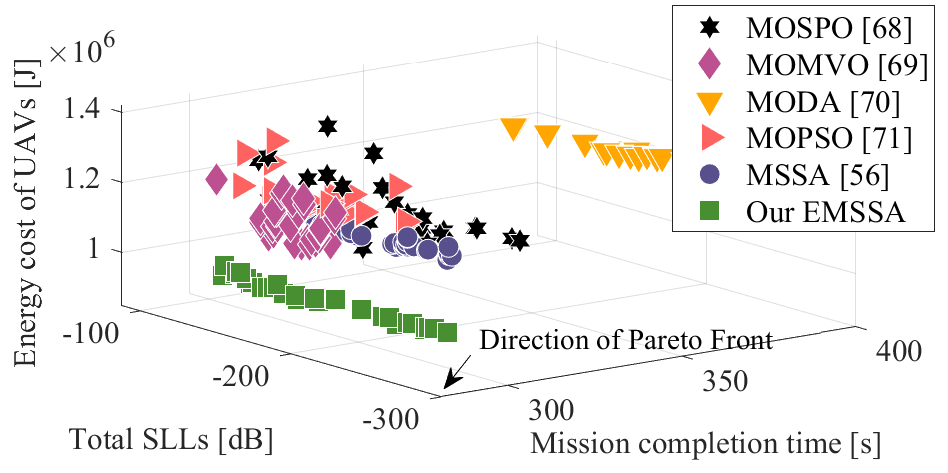}\label{fig:PS-results2}}
    \\
    \centering
      \subfloat[Large-scale scenario under the entire LoS channel.]{
       \includegraphics[width=0.45\linewidth]{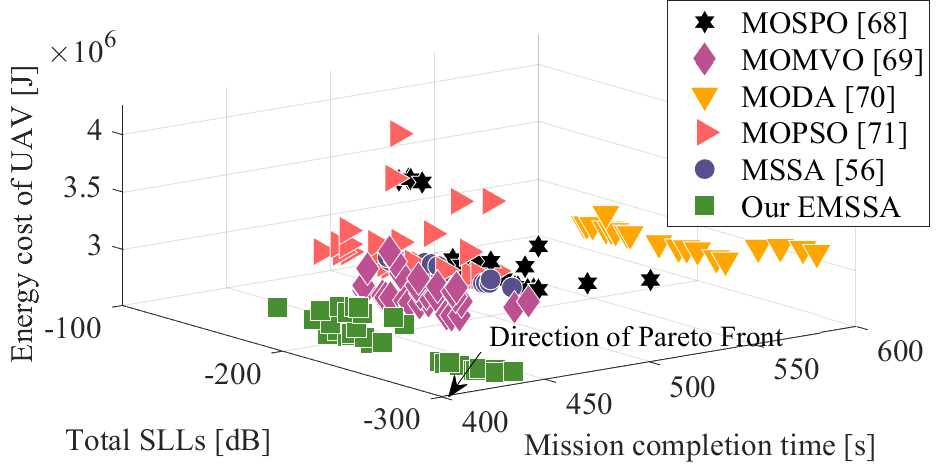}\label{fig:PS-results3}}
    \centering
      \subfloat[Large-scale scenario under the probability LoS channel.]{
       \includegraphics[width=0.45\linewidth]{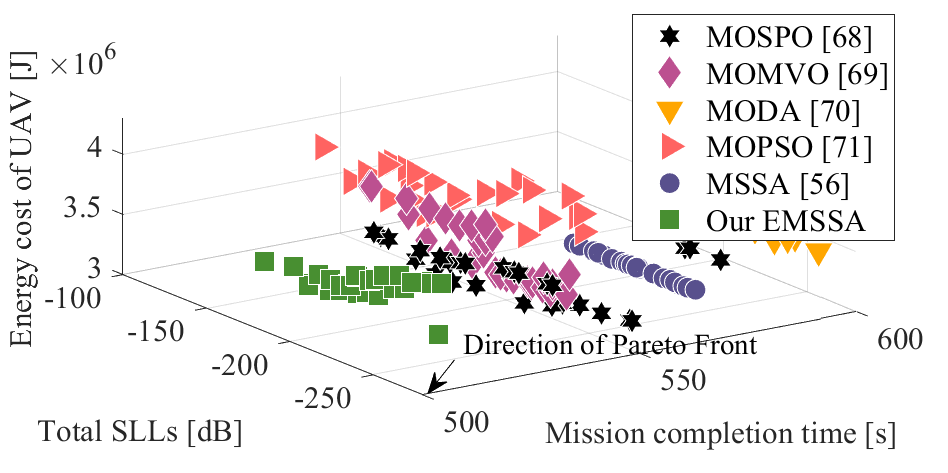}\label{fig:PS-results4}}
	\caption{Pareto solution distributions obtained by different algorithms.}
	\label{fig:PS-results}
\end{figure*}

\begin{table}
	\centering
	\caption{Optimization results obtained by different methods in entire LoS channel condition of small-scale scenario}
	\label{tab:result-los-small}
	\begin{tabular}{llll}
		\toprule
		\bf{Method} 	        & \bf{$f_{1}$ [s] } & \bf{$f_{2}$ [dB] }  & \bf{$f_{3}$ [J] }  \\ \midrule
		RanddomLAA              & $374.07$          & \bm{$-192.34$}           & $9.92 \times 10^{5}$   \\
		MOSPO \cite{Khodadadi2022}           & $246.34$          & $-138.68$           & $1.32 \times 10^{6}$   \\
		MOMVO \cite{Mirjalili2017}           & $247.11$          & $-154.52$           & $1.03 \times 10^{6}$   \\
		MODA  \cite{Mirjalili2016}           & $321.58$          & $-128.39$           & $9.98 \times 10^{5}$   \\
		MOPSO \cite{Coello2002}           & $249.77$          & $-138.38$            & $1.26 \times 10^{6}$   \\
		MSSA \cite{Abualigah2020}            & $269.86$          & $-171.10$           & $8.45 \times 10^{5}$   \\
		Our EMSSA               & \bm{$238.24$}     & $-181.61$      & \bm{$8.43 \times 10^{5}$} \\ \bottomrule
	\end{tabular}
\end{table}

\begin{table}
	\centering
	\caption{Optimization results obtained by different methods in entire LoS channel condition of large-scale scenario}
	\label{tab:result-los-large}
	\begin{tabular}{llll}
		\toprule
		\textbf{Method} 	& \bf{$f_{1}$ [s] } & \bf{$f_{2}$ [dB] } & \bf{$f_{3}$ [J] }  \\ \midrule
		RanddomLAA              & $557.32$          & $-287.88$           & $2.95 \times 10^{6}$   \\
		MOSPO \cite{Khodadadi2022}           & $487.81$          & $-170.56$           & $3.13 \times 10^{6}$   \\
		MOMVO \cite{Mirjalili2017}                  & $448.36$          & $-189.54$           & $3.19 \times 10^{6}$   \\
		MODA \cite{Mirjalili2016}                   & $557.20$          & $-186.12 $          & $3.08 \times 10^{6}$   \\
		MOPSO \cite{Coello2002}                  & $573.29$          & $-218.53$           & $3.27 \times 10^{6}$   \\
		MSSA \cite{Abualigah2020}                   & $465.41$          & $-208.41$           & $3.10 \times 10^{6}$   \\
		Our EMSSA                   & \bm{$439.71$}     & \bm{$-293.07$}       & \bm{$2.58 \times 10^{6}$} \\ \bottomrule
	\end{tabular}
\end{table}

\par In this section, we utilize the proposed EMSSA and other CB-based comparison methods to address the formulated MOP for evaluating the performance of different algorithms. Tables \ref{tab:result-los-small}, \ref{tab:result-los-large}, \ref{tab:result-plos-small} and \ref{tab:result-plos-large} show the numerical results procured by these CB-based methods, including the mission completion time ($f_1$), the total SLL towards the eavesdropper $E$ ($f_2$) and the energy cost of the UAVs ($f_3$) under the entire and probability LoS channel conditions in both small-scale and large-scale scenarios, respectively. It is apparent from the tables that the proposed EMSSA achieves the best performance among almost all optimization objectives in all scale scenarios under the entire LoS channel and the small-scale scenario under the probability LoS channel, which means that the EMSSA outmatches other CB-based methods. Specifically, the optimization results achieved by the EMSSA are much superior to that of the randomLAA method which is with the most probability to be employed in practice, indicating that the considered multi-objective optimization approach is non-trivial and can enhance the efficiency of the data harvesting and dissemination system. The outperformance of the EMSSA means that it is more suitable for solving the formulated MOP compared to other multi-objective optimization algorithms. Moreover, the Pareto solution distributions of these algorithms are presented in Fig. \ref{fig:PS-results}. One can observe that the solutions obtained by EMSSA are more wide-coverage and closer to the direction of the truly Pareto front. Thus, the proposed enhanced measures of the EMSSA are valid and effective for solving the formulated optimization problem.

\begin{table}
	\centering
	\caption{Optimization results obtained by different methods in probability LoS channel condition of small-scale scenario}
	\label{tab:result-plos-small}
	\begin{tabular}{llll}
		\toprule
		\textbf{Method} 	& \bf{$f_{1}$ [s] } & \bf{$f_{2}$ [dB] } & \bf{$f_{3}$ [J] }  \\ \midrule
		RanddomLAA              & $414.90$         & $-192.33$           & $1.10 \times 10^{6}$   \\
		MOSPO \cite{Khodadadi2022}           & $313.27$          & $-107.00$           & $1.26 \times 10^{6}$   \\
		MOMVO \cite{Mirjalili2017}                  & $296.39$         & $-113.93$           & $1.21 \times 10^{6}$   \\
		MODA \cite{Mirjalili2016}                   & $393.10$         & $-87.39$           & $1.17 \times 10^{6}$   \\
		MOPSO \cite{Coello2002}                  & $302.81$         & $-150.23$           & $1.46 \times 10^{6}$   \\
		MSSA \cite{Abualigah2020}                   & $327.82$         & $-165.02$           & $1.14 \times 10^{6}$   \\
		Our EMSSA                   & \bm{$290.98$}    & \bm{$-205.80$}      & \bm{$9.95 \times 10^{5}$} \\ \bottomrule
	\end{tabular}
\end{table}

\begin{table}
	\centering
	\caption{Optimization results obtained by different methods in probability LoS channel condition of large-scale scenario}
	\label{tab:result-plos-large}
	\begin{tabular}{llll}
		\toprule
	    \textbf{Method} 	& \bf{$f_{1}$ [s] } & \bf{$f_{2}$ [dB] } & \bf{$f_{3}$ [J] }  \\ \midrule
		RanddomLAA              & $618.80$         & $-287.87$           & $3.28 \times 10^{6}$   \\
		MOSPO \cite{Khodadadi2022}           & $576.13$          & $-206.21$           & $3.71 \times 10^{6}$   \\
		MOMVO \cite{Mirjalili2017}                  & $532.06$         & $-186.74$           & $3.90 \times 10^{6}$   \\
		MODA \cite{Mirjalili2016}                   & $591.70$         & $-194.02$           & $3.48 \times 10^{6}$   \\
		MOPSO \cite{Coello2002}                  & $531.71$         & $-137.49$           & $4.04 \times 10^{6}$   \\
		MSSA \cite{Abualigah2020}                   & $562.48$         & $-211.31$           & $3.42 \times 10^{6}$   \\
		Our EMSSA                   & \bm{$522.57$}     & \bm{$-311.56$}       & \bm{$3.11 \times 10^{6}$} \\ \bottomrule
	\end{tabular}
\end{table}

%
%
\subsubsection{Comparison with Different Communication Strategies}
\label{ssec:Comparison with Different Communication Strategies}

\begin{table}
	\centering
	\caption{Performance evaluations in entire LoS channel condition of small-scale scenario}
	\label{tab:benchmark-los-small}
	\begin{tabular}{llll}
		\toprule
		\textbf{Benchmarks}             	&  $f_1$ [\textbf{s}]     & $f_3$ [\textbf{J}]     \\ \midrule
		Multi-hop flying ad-hoc network     & $2487.71$         & $2.23 \times 10^6$           \\
		Flying between sensors and BSs         & $1080.28$          & $1.09 \times 10^6$           \\	
		Our CB-based method                     & \bm{$238.24$}     & \bm{$8.43 \times 10^{5}$}    \\ \bottomrule
	\end{tabular}
\end{table}

\begin{table}
	\centering
	\caption{Performance evaluations in entire LoS channel condition of large-scale scenario}
	\label{tab:benchmark-los-large}
	\begin{tabular}{llll}
		\toprule
		\textbf{Benchmarks}             	&  $f_1$ [\textbf{s}]     & $f_3$ [\textbf{J}]     \\ \midrule
		Multi-hop flying ad-hoc network     & $2848.95$         & $6.42 \times 10^6$           \\
		Flying between sensors and BSs         & $1111.42$          & \bm{$2.19 \times 10^6$}           \\	
		Our CB-based method                     & \bm{$439.71$}     & $2.58 \times 10^{6}$    \\ \bottomrule
	\end{tabular}
\end{table}


\begin{table}
	\centering
	\caption{Performance evaluations in probability LoS channel condition of small-scale scenario}
	\label{tab:benchmark-plos-small}
	\begin{tabular}{llll}
		\toprule
		\textbf{Benchmarks}             	&  $f_1$ [\textbf{s}]     & $f_3$ [\textbf{J}]       \\ \midrule
		Multi-hop flying ad-hoc network     & $2500.77$               & $2.26 \times 10^6$       \\
		Flying between sensors and BSs         & $1218.72$          & $1.49 \times 10^6$  \\	
		Our CB-based method                     & \bm{$290.98$}               & \bm{$9.95 \times 10^{5}$}     \\ \bottomrule
	\end{tabular}
\end{table}

\begin{table}
	\centering
	\caption{Performance evaluations in probability LoS channel condition of large-scale scenario}
	\label{tab:benchmark-plos-large}
	\begin{tabular}{llll}
		\toprule
		\textbf{Benchmarks}             	&  $f_1$ [\textbf{s}]     & $f_3$ [\textbf{J}]     \\ \midrule
		Multi-hop flying ad-hoc network     & $2899.87$         & $6.67 \times 10^6$           \\
		Flying between sensors and BSs         & $1201.14$         & \bm{$2.60 \times 10^6$}           \\	
		Our CB-based method                     & \bm{$522.57$}     & $3.11 \times 10^{6}$    \\ \bottomrule
	\end{tabular}
\end{table}

\par We analyze the performance of different communication strategies for the data harvesting and dissemination of IoT via the UAV swarm in this section, and two benchmark strategies that are constructing a multi-hop flying ad-hoc network and flying between IoTs and BSs to disseminate data are introduced, as shown in Fig. \ref{fig:benchmark}. Specifically, in the multi-hop flying ad-hoc network strategy, each IoT cluster is assigned with 8 UAVs to construct a multi-hop flying ad-hoc network to disseminate data from the sensors to these BSs successively. For example, in the small-scale scenario, two sets of UAV swarms with 8 elements will form two multi-hop flying ad-hoc networks simultaneously for serving the two IoT clusters, and then disseminate data from the sensors to these BSs successively. In the flying between IoTs and BSs strategy, the UAVs are dispatched to reach the different IoT clusters for harvesting data, and then fly to different BSs for data dissemination. For instance, in the large-scale scenario, 32 UAVs will disseminate data from 4 IoT clusters to 8 BSs simultaneously. The numerical results of different strategies in both entire and probability LoS channels are given in Tables \ref{tab:benchmark-los-small}, \ref{tab:benchmark-los-large}, \ref{tab:benchmark-plos-small} and \ref{tab:benchmark-plos-large}, respectively. It can be seen from the tables that the proposed CB-based method is more effective than the other two strategies in most cases in terms of time and energy cost. This is because the CB-based method can extend the communication range so that decreasing the flight distance of UAVs. As such, the corresponding flight time and energy can be saved. In contrast, the benchmark strategies require the UAVs to fly close to either the senders or receivers, resulting in a large amount of time and energy cost. However, in a few cases, the CB-based method will consume more energy than the flying between sensors and BSs scheme. This is due to the fact that some UAVs must fly to their locations at a faster speed in order for multiple UAVs to reach their designated locations simultaneously, resulting in more energy costs of UAVs. Clearly, this case is especially likely to occur in large-scale scenarios where large amounts of UAV speeds are controlled at the same time. Thus, our approach is more suitable for scenarios that are time sensitive and have a moderate number of UAVs.

%
%
\subsection{Performance Verifications under Special Cases}
\label{sec:Special Cases}

\par In this part, we verify the performance of our proposed method under some special cases, including CSI error, UAV jitter, phase synchronization error, and using narrow-band IoT cases. They are provided in Appendix C of the supplemental material.

\section{Conclusion} 
\label{sec:conclusion}

\par In this paper, the CB-based data harvesting and dissemination via the sensor-enabled GVAA and UAV-enabled AVAA in IoT were investigated. We first considered a typical scenario that several IoT clusters need to transmit the sensed data to the remote BSs, in which a UAV swarm and an eavesdropper exist for assisting terrestrial networks and intercepting the communications, respectively. Then, we formulated an MOP to optimize the total mission completion time, the SLL towards the eavesdropper, and the energy cost of the UAV swarm. The formulated optimization problem was proven to be an NP-hard, mixed-variable, and large-scale problem. Thus, we proposed an EMSSA by enhancing the solution initialization, algorithm parameter update, and solution update of the conventional MSSA to make it more suitable for dealing with formulated MOP. The algorithm can find a set of candidate solutions with different trade-offs which can meet various requirements in a low computational complexity. Simulation results demonstrated the proposed EMSSA outmatches other MOP algorithms and also showed that the proposed method can reduce time and energy costs significantly compared with some benchmark strategies which require the UAVs frequently fly (such as the UAVs construct a multi-hop link or directly fly between IoTs and access points). We also found that the proposed method is also valid by considering unexpected circumstances. The results of this paper can be further extended by considering other channel models or practice experiments, which will be investigated as future work.


\bibliographystyle{IEEEtran}
\bibliography{myref}

\begin{IEEEbiography}[{\includegraphics[width=1in,height=1.25in,clip,keepaspectratio]{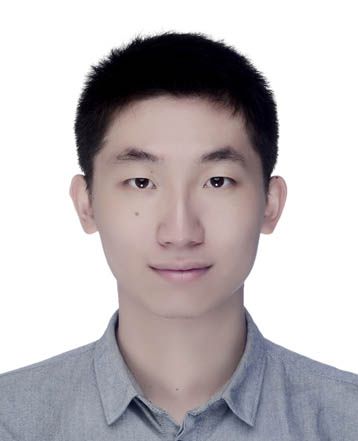}}]{Jiahui Li}
(S'21) received a BS degree in Software Engineering, and an MS degree in Computer Science and Technology from Jilin University, Changchun, China, in 2018 and 2021, respectively. He is currently studying Computer Science at Jilin University to get a Ph.D. degree, and also a visiting Ph. D. at Singapore University of Technology and Design (SUTD), Singapore. His current research focuses on UAV networks, antenna arrays, and optimization.
\end{IEEEbiography}

\begin{IEEEbiography}[{\includegraphics[width=1in,height=1.25in,clip,keepaspectratio]{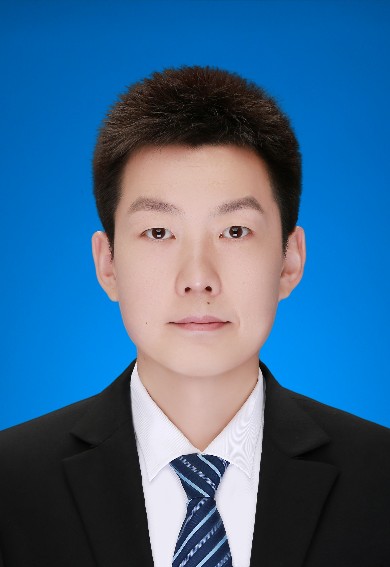}}]{Geng Sun}
(S'17-M'19) received the B.S. degree in communication engineering from Dalian Polytechnic University, and the Ph.D. degree in computer science and technology from Jilin University, in 2011 and 2018, respectively. He was a Visiting Researcher with the School of Electrical and Computer Engineering, Georgia Institute of Technology, USA. He is an Associate Professor in College of Computer Science and Technology at Jilin University, and His research interests include wireless networks, UAV communications, collaborative beamforming and optimizations.
\end{IEEEbiography}

\begin{IEEEbiography}[{\includegraphics[width=1in,height=1.25in,clip,keepaspectratio]{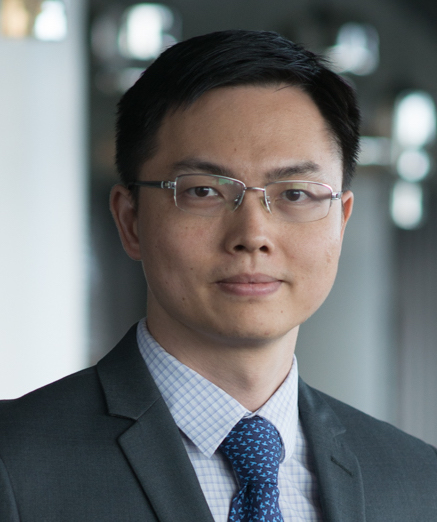}}]{Lingjie Duan}
(S'09-M'12-SM'17) received the Ph.D. degree from The Chinese University of Hong Kong in 2012.  He is an Associate Professor of Engineering Systems and Design with the Singapore University of Technology and Design (SUTD). In 2011, he was a Visiting Scholar at University of California at Berkeley, Berkeley, CA, USA. His research interests include network economics and game theory, cognitive and green networks, and energy harvesting wireless communications. He is an Editor of IEEE Transactions on Wireless Communications. He was an Editor of IEEE Communications Surveys and Tutorials. He also served as a Guest Editor of the IEEE Journal on Selected Areas in Communications Special Issue on Human-in-the-Loop Mobile Networks, as well as IEEE Wireless Communications Magazine. He received the SUTD Excellence in Research Award in 2016 and the 10th IEEE ComSoc AsiaPacific Outstanding Young Researcher Award in 2015.
\end{IEEEbiography}

\begin{IEEEbiography}[{\includegraphics[width=1in,height=1.25in,clip,keepaspectratio]{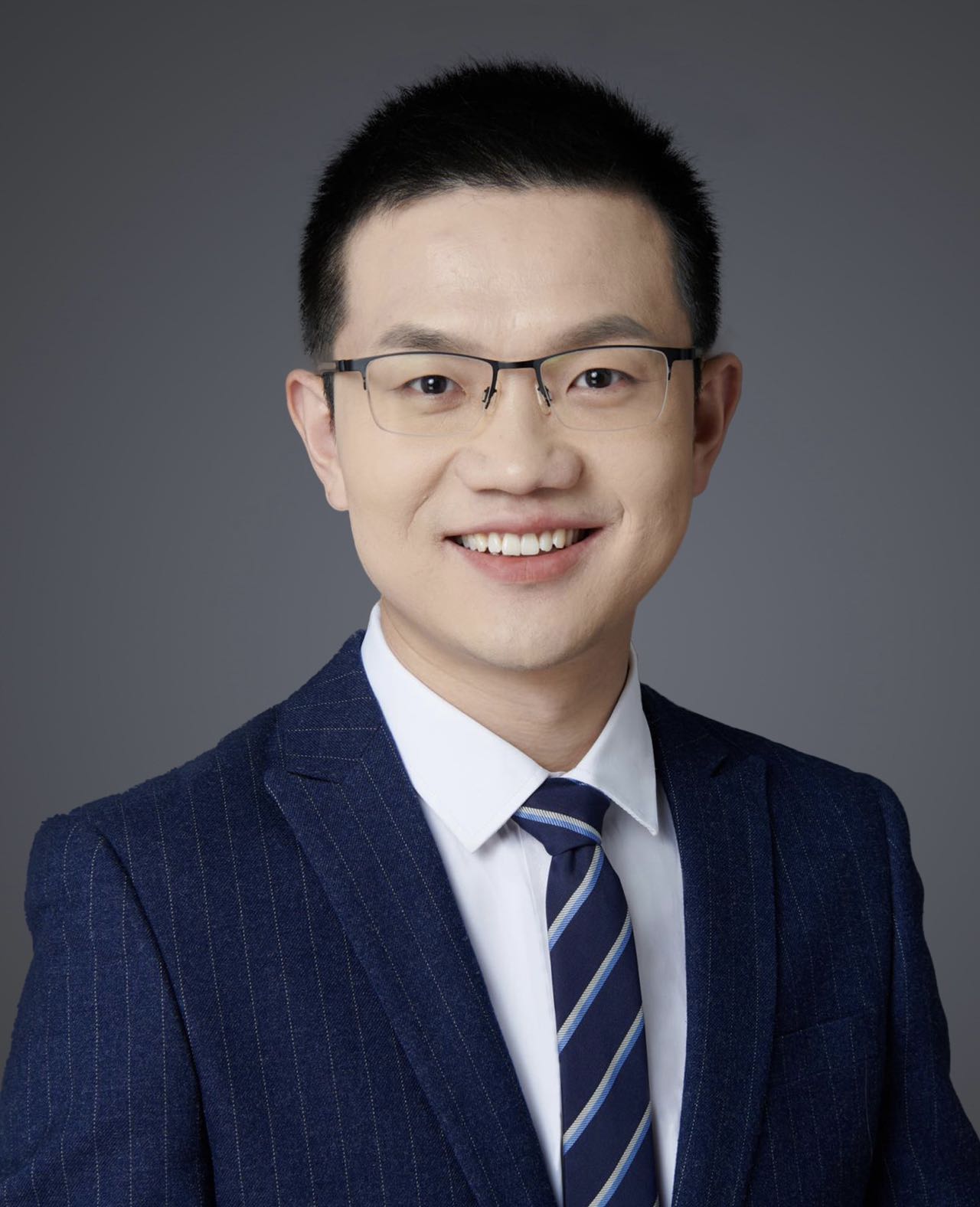}}]{Qingqing Wu}
(S'13-M'16-SM'22) received the B.Eng. and the Ph.D. degrees in Electronic Engineering from South China University of Technology and Shanghai Jiao Tong University (SJTU) in 2012 and 2016, respectively. From 2016 to 2020, he was a Research Fellow in the Department of Electrical and Computer Engineering at National University of Singapore. He is currently an Associate Professor with Shanghai Jiao Tong University. His current research interest includes intelligent reflecting surface (IRS), unmanned aerial vehicle (UAV) communications, and MIMO transceiver design. He has coauthored more than 100 IEEE journal papers with 26 ESI highly cited papers and 8 ESI hot papers, which have received more than 18,000 Google citations. He was listed as the Clarivate ESI Highly Cited Researcher in 2022 and 2021, the Most Influential Scholar Award in AI-2000 by Aminer in 2021 and World’s Top 2\% Scientist by Stanford University in 2020 and 2021.

\end{IEEEbiography}

\end{document}